\documentclass{article}
\usepackage{ijcai16}
\usepackage{times}

\usepackage{graphicx} 
\usepackage{subfigure}
\usepackage{caption}

\newcommand{\figwidthtwo}{0.48\textwidth}
\newcommand{\figwidththree}{0.32\textwidth}
\newcommand{\figwidthfour}{0.235\textwidth}

\usepackage{amsmath,amsthm,amsfonts,amssymb}

\usepackage{algorithm}
\usepackage{algpseudocode}

\usepackage{enumitem}

\include{Definitions}

\newcommand{\tLSTD}{\text{t-LSTD}}

\newcommand{\Actions}{\mathcal{A}}
\newcommand{\States}{\mathcal{S}}
\newcommand{\Pfcn}{\mathrm{Pr}}

\newcommand{\Rfcn}{r}

\newcommand{\Amat}{\mathbf{A}}

\newcommand{\Dmat}{\mathbf{D}}
\newcommand{\Kmat}{\mathbf{K}}
\newcommand{\Lmat}{\mathbf{L}}

\newcommand{\Ppi}{\mathbf{P}^\pi}
\newcommand{\Qmat}{\mathbf{Q}}
\newcommand{\Rmat}{\mathbf{R}}
\newcommand{\Umat}{\mathbf{U}}
\newcommand{\Vmat}{\mathbf{V}}
\newcommand{\Xmat}{\mathbf{X}}
\newcommand{\Zmat}{\mathbf{Z}}

\newcommand{\Sigmamat}{\boldsymbol{\Sigma}}

\newcommand{\Lmatnext}{\tilde{\Lmat}}
\newcommand{\Rmatnext}{\tilde{\Rmat}}

\newcommand{\bvec}{\mathbf{b}}
\newcommand{\dvec}{\mathbf{d}}

\newcommand{\mvec}{\mathbf{m}}
\newcommand{\nvec}{\mathbf{n}}
\newcommand{\pvec}{\mathbf{p}}
\newcommand{\qvec}{\mathbf{q}}

\newcommand{\uvec}{\mathbf{u}}
\newcommand{\vvec}{\mathbf{v}}
\newcommand{\wvec}{\mathbf{w}}
\newcommand{\xvec}{\mathbf{x}}
\newcommand{\zvec}{\mathbf{z}}

\newcommand{\rpi}{\mathbf{r}_{\pi}}

\newcommand{\xdim}{d}
\newcommand{\rdim}{r}
\newcommand{\nsamples}{T}
\newcommand{\nstates}{n}
\newcommand{\rankA}{{\text{rank}(\Amat)}}
\newcommand{\picardpower}{p}

\newcommand{\updelay}{k}

\newcommand{\ralignspace}{\vspace{-0.15cm}}
\newcommand{\balignspace}{\par\vspace{-0.2cm}}

\newcommand{\citep}[1]{\cite{#1}}
\newcommand{\citet}[1]{\citeauthor{#1} [\citeyear{#1}]}

\begin{document}

\title{Incremental Truncated LSTD}
\author{Clement Gehring \\ MIT CSAIL\\ Cambridge MA 02139 USA\\
gehring@csail.mit.edu
\And
Yangchen Pan  \and Martha White\\
Indiana University\\ Bloomington IN 47405 USA\\
\{yangpan, martha\}@indiana.edu}

\maketitle

\begin{abstract}
Balancing between computational efficiency and 
sample efficiency is an important goal in
reinforcement learning.
Temporal difference (TD) learning algorithms stochastically update
the value function, with a linear time complexity in the number of features, whereas
least-squares temporal difference (LSTD) algorithms
are sample efficient but can be quadratic in the number of features. 
In this work, we develop
an efficient incremental low-rank LSTD($\lambda$) algorithm
that progresses towards the goal of better balancing computation and sample efficiency. 
The algorithm reduces the computation and storage
complexity to the number of features times the chosen rank parameter
while summarizing past samples efficiently to nearly obtain the sample efficiency of LSTD. 
We derive a simulation bound on the solution given by truncated low-rank approximation,
illustrating a bias-variance trade-off dependent on the choice of rank. 
We demonstrate 
that the algorithm effectively balances computational complexity
and sample efficiency
for policy evaluation in a benchmark task
and a high-dimensional energy allocation domain. 
\end{abstract}

\section{Introduction}
Value function approximation is a central goal in reinforcement learning.
A common approach to learn the value function is to minimize
the mean-squared projected Bellman error, with dominant approaches
generally split into stochastic temporal difference (TD) methods
and least squares temporal difference (LSTD) methods. 
TD learning \citep{sutton1988learning}
requires only $O(\xdim)$ computation and storage per step for $\xdim$ features,
but can be sample inefficient \citep{bradtke1996linear,boyan1999least,geramifard2006incremental}
because a sample is used only once for a stochastic update.
Nonetheless, for practical incremental updating, particularly for high-dimensional features, 
it remains a dominant approach. 

On the other end of the spectrum, 
LSTD \citep{bradtke1996linear} algorithms summarizes all past data into a linear system,
and are more sample efficient than TD \citep{bradtke1996linear,boyan1999least,geramifard2006incremental},
but at the cost of higher computational complexity and storage complexity. 
Several algorithms have been proposed to tackle these practical 
issues,\footnote{A somewhat orthogonal strategy is to sub-select features before applying LSTD \citep{keller2006automatic}.
Feature selection is an important topic on its own;
we therefore focus exploration on direct approximations of the LSTD system itself.}
including 
iLSTD \citep{geramifard2006incremental},
iLSTD($\lambda$) \citep{geramifard2007ilstd},
sigma-point policy iteration \citep{bowling2008sigma},
random projections \citep{ghavamzadeh2010lstd},
experience replay strategies \citep{lin1993reinforcement,prashanth2013fast} 
and 
forgetful LSTD \citep{vanseijen2015adeeper}. 
Practical incremental LSTD strategies typically consist of using the system as a model \citep{geramifard2006incremental,geramifard2007ilstd,bowling2008sigma}, similar to experience replay,
or using random projections to reduce the size of the system \citep{ghavamzadeh2010lstd}. 
To date, however,
none seem to take advantage of the fact that the LSTD system is likely low-rank,
due to dependent features \citep{bertsekas2007dynamic}, 
small numbers of samples \citep{kolter2009regularization,ghavamzadeh2010lstd}
or principal subspaces or highways in the environment \citep{keller2006automatic}.

 In this work, we propose \tLSTD, a novel incremental low-rank LSTD($\lambda$),
 to further bridge the gap between computation and sample efficiency.
 The key advantage to using a low-rank approximation is to direct approximation
 to less significant parts of the system. 
 For the original linear system 
 with $\xdim$ features and corresponding $\xdim \times \xdim$ matrix,
 we incrementally maintain a truncated rank $\rdim$ singular value decomposition (SVD),
 which reduces storage to significantly smaller $\xdim \times \rdim$ matrices
 and computation to $O(\xdim \rdim + r^3)$.
 In addition to these practical computational gains,
 this approach has several key benefits. 
 First, it exploits the fact that the linear system likely has redundancies,
 reducing computation and storage
 without sacrificing much accuracy. 
 Second, the resulting solution is better conditioned, as truncating
 singular values is a form of regularization. Regularization strategies have proven effective for stability
 \citep{bertsekas2007dynamic,farahmand2008regularized,kolter2009regularization,farahmand2011thesis};
 however, unlike these previous approaches, the truncated SVD also 
 reduces the size of the system.
 Third, like iLSTD, it provides a close approximation to the system,
  but with storage complexity reduced to O($\xdim \rdim$) instead of O($\xdim^2$)
  and a more intuitive toggle $\rdim$ to balance computation and approximation. 
Finally, the approach is more promising for tracking and for control, because previous
samples can be efficiently down-weighted in O($\rdim$)
and
the solution can be computed in O($\xdim \rdim$) time, enabling every-step updating.
 
 To better investigate the merit of low-rank approximations for LSTD,
 we first derive a new simulation bound for low-rank approximations, highlighting
 the bias-variance trade-off given by this form of regularization. 
 We then 
 empirically investigate the rank properties of the system in a benchmark task (Mountain Car)
 with common feature representations (tile coding and RBFs),
  to explore the validity of using low-rank approximation in reinforcement learning.
  Finally, we demonstrate efficacy of \tLSTD\ for value function approximation
  in this domain as well as a high-dimensional energy allocation domain.

\section{Problem formulation}

We assume the agent interacts with and receives reward from an environment
formalized by a Markov decision process: $(\States, \Actions, \Pfcn, \Rfcn,\gamma)$
where 
$\States$ is the set of states, $\nstates = | \States |$;
$\Actions$ is the set of actions;
$\Pfcn: \States \times \Actions \times \States \rightarrow [0,1]$ is the transition probability function;
$\Rfcn: \States \times \Actions \times \States \rightarrow \RR$ is the reward function,
where
$\Pfcn(s,a,s')$ is the probability of transitioning from state $s$ into state $s'$ when taking action $a$,
receiving reward $\Rfcn(s,a,s')$;
and $\gamma \in [0,1]$ is the discount rate.
For a policy $\pi: \States \times \Actions \rightarrow [0,1]$, where $\sum_{a\in \Actions} \pi(s,a) = 1 \; \forall s \in \States$,
define matrix
$\Ppi\in \RR^{\nstates \times \nstates}$ as $\Ppi(s,s') = \sum_{a \in \Actions} \pi(s,a) \Pfcn(s, a, s')$
and vector $\rpi \in \RR^\nstates$ as the average rewards from each state under $\pi$. 
The value at a state $s_t$
is the expected discounted sum of future rewards, assuming actions are selected according to $\pi$,
%
\ralignspace
\begin{align*}
V^\pi(s_t) 
= \rpi(s_t) + \gamma \sum_{s_{t+1} \in \States} \Ppi(s_t, s_{t+1}) V^\pi(s_{t+1})
.
\end{align*}
\balignspace
\noindent
Value function learning
using linear function approximation
can be expressed as a linear system \citep{bradtke1996linear}: 
$\Amat \wvec = \bvec$ for
\ralignspace
\begin{align*}
\Amat &= \Xmat^\top \Dmat (\eye - \gamma \lambda \Ppi)^\inv (\eye - \gamma \Ppi)  \Xmat \\
\bvec &= \Xmat^\top \Dmat (\eye - \gamma \lambda \Ppi)^\inv \rpi
\end{align*}
\balignspace
\noindent
where each row in $\Xmat \in \RR^{\nstates \times \xdim}$ corresponds to the features for a state; 
$\Dmat$ is a diagonal matrix with the stationary distribution of $\pi$ on the diagonal;
and $\lambda$ is the trace parameter for the $\lambda$-return.
For action-value function approximation, the system is the same, but with state-action
features in $\Xmat$. 
These matrices are approximated using  
\ralignspace
\begin{align*}
\Amat_\nsamples = \frac{1}{\nsamples} \sum_{t=0}^{\nsamples-1} \zvec_t (\xvec_t - \gamma \xvec_{t+1})^\top
\hspace{0.2cm}\text{and}\hspace{0.3cm}
\bvec_\nsamples = \frac{1}{\nsamples} \sum_{t=0}^{\nsamples-1} \zvec_t r_{t+1}
\end{align*}
\balignspace
\noindent
for eligibility trace $\zvec_t = \sum_{i=0}^t (\gamma\lambda)^{t-i} \xvec_i$
and sampled trajectory $s_0, a_0, r_1, s_1, a_1, \ldots, s_{\nsamples-1}, a_{\nsamples-1}, r_\nsamples, s_\nsamples$.

There are several strategies to solve this system incrementally.
A standard approach is to use TD and variants, which stochastically update $\wvec$ with new samples
as $\wvec = \wvec + \alpha (r_{t+1} + \gamma \xvec_{t+1}^\top \wvec - \xvec_{t}^\top \wvec) \zvec_t$.
The LSTD algorithms instead incrementally approximate these matrices or corresponding system.
For example, the original LSTD algorithm \citep{bradtke1996linear} incrementally maintains $\Amat_t^\inv$ using the matrix inversion lemma
so that on each step the new solution $\wvec = \Amat_t^\inv \bvec_t$ can be computed.

We iteratively update and solve this system
by maintaining a low rank approximation to $\Amat_t$ directly.
Any matrix $\Amat \in \RR^{\xdim \times \xdim}$ has a singular value decomposition (SVD)
$\Amat = \Umat \Sigmamat \Vmat^\top$, where $\Sigmamat \in \RR^{\xdim \times \xdim}$ 
is a diagonal matrix of 
the singular values of $\Amat$ and $\Umat, \Vmat \in \RR^{\xdim \times \xdim}$ are
orthonormal matrices: $\Umat^\top \Umat = \eye = \Vmat^\top \Vmat$ and $\Umat \Umat^\top = \eye = \Vmat \Vmat^\top$. 
With this decomposition, for full rank $\Amat$, the inverse of $\Amat$ is simply computed by inverting the singular values,
to get $\wvec = \Amat^\inv \bvec = \Vmat \Sigmamat^\inv \Umat^\top \bvec$.
In many cases, however, the rank of $\Amat$ may be smaller than $\xdim$,
giving $\xdim - \rankA$ singular values that are zero. Further, we
can approximate $\Amat$ by dropping (i.e., zeroing) some number of the smallest
singular values, to obtain a rank $\rdim$ approximation. 
Correspondingly,
rows of $\Umat$ and $\Vmat$ are zeroed,
reducing the size of these matrices to $\xdim \times \rdim$.
The further we reduce the dimension, the more practical for efficient
incremental updating; however there is clearly a trade-off in terms of accuracy of the solution. 
We first investigate the theoretical properties of using 
a low-rank approximation to $\Amat_t$ and then present
the incremental \tLSTD\ algorithm.

\section{Characterizing the low-rank approximation}

Low-rank approximations provide an efficient 
approach to obtaining stable solutions for linear systems. 
The approach is particularly well motivated for our resource constrained setting,
because of the classical Eckart-Young-Mirsky theorem \citep{eckhart1936the,mirsky1960symmetric},
which states that
the optimal rank $\rdim$ approximation to a matrix under any unitarily invariant norm (e.g., Frobenius norm, spectral norm, nuclear norm)
is the truncated singular value decomposition. 
In addition to this nice property, which facilitates development of an efficient approximate LSTD algorithm,
the truncated SVD can be viewed as a form of regularization \citep{hansen1986thetruncated}, improving
the stability of the solution. 

To see why truncated SVD regularizes the solution, consider the solution to the linear system
\ralignspace
\begin{align*}
\wvec = \Amat^\pinv \bvec = \Vmat \Sigmamat^\pinv \Umat^\top \bvec 
= \sum_{i=1}^{\rankA} \frac{\vvec_i \uvec_i^\top}{\sigma_i} \bvec
\end{align*}
\balignspace
\noindent
for ordered singular values $\sigma_1 \ge \sigma_2 \ge \ldots \sigma_{\rankA} > \sigma_{\rankA+1} = 0,  \ldots, \sigma_\xdim = 0$.
$\Amat^\pinv$ is the pseudo-inverse of $\Amat$, 
with $\Sigmamat^\pinv =  \diag(\sigma_1^\inv \!\!, ..., \sigma_{\rankA}^\inv  \!\!, 0, ..., \!0)$
composed of the inverses of the non-zero singular values.  
For very small, but still non-zero $\sigma_i$, the outer product $\vvec_i \uvec_i^\top$ will be scaled by a large number;
this will often correspond to highly overfitting the observed samples and a high variance estimate. 
A common practice is to regularize $\wvec$
with $\eta \| \wvec \|_2$ for regularization weight 
$\eta \!>\! 0$, 
modifying the multiplier from $\sigma_i^\inv$ to $\sigma_i/(\sigma_i^2 + \eta)$ because
$\wvec = (\Amat^\top \Amat + \eta \eye)^\inv \Amat^\top \bvec = \Vmat (\Sigmamat^2 + \eta \eye)^\inv \Sigmamat \Umat^\top \bvec$. 
The regularization reduces variance but introduces bias controlled by $\eta$;
for $\eta = 0$, we obtain the unbiased solution. 
Similarly, by thresholding the smallest singular values to retain only the top $\rdim$ singular values,  
\ralignspace
\begin{align*}
\wvec \!=\! \Amat_\rdim^\pinv \bvec \!=\!  \Vmat  \! \diag(\sigma_1^\inv \!\!, ..., \sigma_\rdim^\inv  \!\!, 0, ..., \!0) \Umat^\top \bvec 
\!=\!\! \sum_{i=1}^{\rdim} \frac{\vvec_i \uvec_i^\top}{\sigma_i} \bvec
\end{align*}
\balignspace
\noindent
we 
introduce bias, but reduce variance
because the size of $\sigma_\rdim^\inv$ can be controlled by the choice of $\rdim < \rankA$. 

To characterize the bias-variance tradeoff, 
we bound the difference between the true solution, $\wvec^*$,
and the approximate
rank $\rdim$ solution at time $t$, $\wvec_{t,r}$.
We use a similar analysis to the one used for regularized LSTD \cite[Proposition 6.3.4]{bertsekas2007dynamic}.
%
This previous bound does not easily extend, because in regularized LSTD, the singular values
are scaled up, maintaining the information in the singular vectors (i.e., no columns are dropped from $\Umat$ or $\Vmat$).
We bound the loss incurred by dropping singular vectors using
insights from work on ill-posed systems. 

The following is a simple but realistic assumption for ill-posed systems \citep{hansen1990thediscrete}.
The assumption states that $\uvec_i^\top \bvec$ shrinks faster than $\sigma_i^\picardpower$,
where $\picardpower$ specifies the smoothness of the solution $\wvec$ and is related
to the smoothness parameter for the Hilbert space setting \cite[Cor. 1.2.7]{groetsch1984thetheory}.

%
\noindent
\textbf{Assumption 1: } The linear system defined by $\Amat = \Umat \Sigmamat \Vmat^\top$ and $\bvec$ satisfy the \textit{discrete Picard condition}:
for some $\picardpower > 1$, 
\begin{align*}
|\uvec_i^\top \bvec | &\le \sigma_i^\picardpower &&\text{for } i = 1, \ldots, \rank(\Amat)\\
|\uvec_i^\top \bvec | &\le \sigma_{\rank(A)}^\picardpower &&\text{for } i = \rank(\Amat)+1, \ldots, \xdim
.
\end{align*}

\noindent
\textbf{Assumption 2:}
As $t \rightarrow \infty$, the sample average $\Amat_t$ converges to the true $\Amat$.
This assumption  
can be satisfied with typical technical assumptions (see \cite{tsitsiklis1997ananalysis}).

We write the SVD of $\Amat = \Umat \Sigmamat \Vmat^\top$ 
and $\Amat_{t} =  \hat{\Umat} \hat{\Sigmamat} \hat{\Vmat}^\top$,
where to avoid cluttered notation, we do not explicitly subscript with $t$.
Further, though the singular values are unique, there is a space of
equivalent singular vectors, up to sign changes 
and multiplication by rotation matrices.
We assume that among the space of equivalent SVDs of $\Amat_{t}$,
the most similar singular vectors for each singular value are chosen between $\Amat$ and $\Amat_{t}$.
This avoids uniqueness issues without losing generality, because we only
conceptually compare the
SVDs of $\Amat$ and $\Amat_{t}$; the proof does not rely on practically obtaining
this matching SVD. 


\newcommand{\sighat}[2]{\hat{\sigma}_{#2}}

\begin{theorem}[Bias-variance trade-off of rank-$\rdim$ approximation]\label{thm_tradeoff}
Let $\Amat_{t,\rdim} =  \hat{\Umat} \hat{\Sigmamat}_{\rdim} \hat{\Vmat}^\top$ be the approximated $\Amat$
after $t$ samples, truncated to rank $\rdim$, i.e., with the last $\rdim+1, \ldots, \xdim$ singular values zeroed.
Let $\wvec^* = \Amat^\pinv \bvec$ and $\wvec_{t,\rdim} = \Amat_{t,\rdim}^\pinv \bvec_t$.
Under Assumption 1 and 2, the relative error of the rank-$\rdim$ weights to the true weights $\wvec^*$ is bounded as follows:
\ralignspace
\begin{align*}
\| \wvec_{t,\rdim} - \wvec^* \|_2
  &\le
   \frac{1}{\sighat{t}{\rdim}} \| \bvec_t - \Amat_t \wvec^*  \|_2 
   + 
 (\xdim - \rdim)  \epsilon(t)\\
    &+
 \underbrace{ (\xdim - \rdim) \sigma_{\rdim}^{\picardpower - 1}}_{\text{bias}}
\end{align*}
\balignspace
for function 
$\epsilon: \NN \rightarrow [0, \infty)$, 
where
$\epsilon(t) \rightarrow 0$ as $t \rightarrow \infty$:
{\small
\begin{align*}
\epsilon(t) = \min \Big(
   &\rankA  \sigma_{1}^{\picardpower-1}, \\
   &\sum_{j=1}^\rankA \sqnorm{\vvec_j \sigma_j^{p-1} - \hat{\vvec}_j \sighat_j^{p-1}} + \sighat_\rdim^{p-1} - \sigma_\rdim^{p-1}\Big)
.
\end{align*}
}
\end{theorem}
A detailed proof is provided in an appendix,
and will be posted with the paper. 
The key step is to split up the error into two terms: 
approximation error due to a finite number of samples $t$
and
bias due the choice of $\rdim < \xdim$.
Then the second part is bounded using the discrete Picard condition
to ensure that the magnitude of $\uvec_j^\top \bvec$ 
does not dominate the error, and by adding and subtracting terms
to express the error in terms of differences between $\Amat$ and $\Amat_t$.
Because $\Amat_t$ converges to $\Amat$, we can see that $\epsilon(t)$
converges to zero because the differences
$\vvec_j \sigma_j^{p-1} - \hat{\vvec}_j \sighat_j^{p-1}$ and $\sighat_\rdim^{p-1} - \sigma_\rdim^{p-1}$
converge to zero.

\vspace{0.1cm}
\noindent
\textbf{Remark 1:} Notice that for no truncation, the bias term disappears and the first term could be very large
because $\hat{\sigma}_\rdim = \hat{\sigma}_\xdim$ could be very small (and often is for systems studied to-date, including in the below experiments). 
In fact, previous work on finite sample analysis of LSTD 
uses an unbiased estimate and the bound suffers from an inverse relationship to the smallest
eigenvalue of $\Xmat^\top \Xmat$ (see \cite[Lemma 3]{lazaric2010finite}, \citep{ghavamzadeh2010lstd,tagorti2015ontherate}). 
Here, we avoid such a potentially large
constant in the bound at the expense of an additional bias term determined by the choice of $\rdim$.
Lasso-TD \citep{ghavamzadeh2011finite} similarly avoids such a dependence, using $\ell_1$ regularization;
to the best of our knowledge, however, there does not yet exist an efficient incremental Lasso-TD algorithm.
A future goal is to use the above bound, to obtain a finite sample bound for \tLSTD($\lambda$),
using the most up-to-date analysis by \citet{tagorti2015ontherate}
and more general techniques for linear system introduced by \citet{pires2012statistical}. 


\vspace{0.1cm}
\noindent
\textbf{Remark 2:}
The discrete Picard condition could be relaxed to an average discrete Picard condition,
where $| \uvec_i ^\top \bvec|$ on average is similar to $\sigma_i^\inv$,
with a bound on the variance of this ratio. 
The assumption above, however, simplifies
the analysis and much more clearly illustrates the importance
of the decay of $\uvec_i^\top \bvec$ for obtaining stable LSTD solutions.

\section{Incremental low-rank LSTD($\lambda$) algorithm}

We have shown that a low-rank approximation to $\Amat_t$ is effective for computing
the solution to LSTD from $t$ samples. However, the computational complexity of 
explicitly computing $\Amat_t$ from samples and then performing a SVD
is $O(\xdim^3)$, which is not feasible for most settings. 
In this section, we propose an algorithm that incrementally computes
a low-rank singular value decomposition of $\Amat_t$, from samples,
with significantly improved storage O($\xdim \rdim$) and computational complexity 
$O(\xdim \rdim + \rdim^3)$, which we can further reduced to O($\xdim \rdim$) 
using mini-batches of size $\rdim$.

\begin{algorithm}[htp!]
\caption{\tLSTD($\lambda$) using incremental SVD}
\label{alg_tlstd}
\begin{algorithmic}
\State // Input rank $\rdim$, and mini-batch size $\updelay$
\State //  with differing update-svd for $\updelay =1$ and $\updelay > 1$
\State $\Umat \gets [], \Vmat \gets [], \Sigmamat \gets 0, \bvec \gets \zerovec, \zvec \gets \zerovec, i \gets 0, t \gets 1$ 
  \State  $\xvec \gets$ the initial observation 
\Repeat
\State Take action according to $\pi$, observe $\xvec'$, reward $r$
\State $\beta \gets 1/(t + \updelay)$
\State $ \zvec\gets \gamma\lambda \zvec + \xvec$
\State $ \dvec\gets \beta(\xvec - \gamma \xvec')$
\State $\Zmat_{:,i} \gets \zvec$  
\State $\Dmat_{:,i} \gets \dvec$
\State $ \bvec \gets (1-\beta) \bvec + \beta \zvec r$
\State $ i \gets i + 1$
\If {$i\geq \updelay$}
\State // Returns $\Umat, \Vmat \in \RR^{\xdim \times \rdim}$, diagonal $\Sigmamat \in \RR^{\rdim \times \rdim}$ 
	\State $\Umat,\Sigmamat , \Vmat \gets$
	\State \text{update-svd}($\Umat,(1-\beta)\Sigmamat ,\Vmat, \sqrt{\beta} \Zmat, \sqrt{\beta} \Dmat, \rdim$)
    \State $\Zmat\gets 0^{\xdim\times k}, \Dmat\gets 0^{\xdim\times k}, i\gets 0, t\gets t + \updelay$
\EndIf
\State $\wvec \gets \Vmat \Sigmamat^\pinv \Umat^\top \bvec$ // $O(\xdim \rdim)$ time
\Until{agent done interaction with environment}
\end{algorithmic}
\end{algorithm}

To maintain a low-rank approximation to $\Amat_t$ incrementally, 
we need to update the SVD with new samples. 
With each new $\xvec_t$, we add the rank-one matrix 
$\zvec_t (\xvec_t - \gamma \xvec_{t+1})^\top$ to $\Amat_t$.
Consequently, we can take advantage of
recent advances for fast low-rank SVD updates \citep{brand2006fast},
with some specialized computational improvements for our setting.
Algorithm \ref{alg_tlstd} summarizes the generic incremental update
for \tLSTD, which can use mini-batches or update on each step, depending
on the choice of the mini-batch size $k$.
Due to space constraints, the detailed pseudo-code for the SVD updates are left out but detailed code and explanations will be published on-line. The basics of the SVD update follow from previous work \citep{brand2006fast} but our implementation offers some optimizations specific for the LSTD case.

By maintaining the SVD incrementally, we do not need to
explicitly maintain $\Amat_t$; therefore, storage is reduced 
to the size of the truncated singular vector matrices, which is O($\xdim \rdim$).
To maintain $O(\xdim \rdim)$ computational complexity, matrix and vector multiplications
need to be carefully ordered. For example, to compute $\wvec$, 
first $\tilde{\bvec} = \Umat^\top \bvec$ is computed in $O(\xdim \rdim)$, 
then $\Sigmamat_\rdim \tilde{\bvec}$ is computed in $O(\rdim)$,
and finally that is multiplied by $\Vmat$ in $O(\xdim \rdim)$.
For $k = 1$ (update on each step), 
the $O(\rdim^3)$ computation arises from a re-diagonalization and the multiplication of 
the resulting orthonormal matrices.
For mini-batches of size $k = \rdim$,
we can get further computational improvements 
by amortizing costs across $\rdim$ steps, 
to obtain a total amortized complexity $O(\xdim \rdim)$, losing the $r^3$ term.

As an additional benefit, unlike previous incremental LSTD algorithms,
we maintain normalized $\Amat_t$ and $\bvec_t$, by incorporating
the term $\beta$. On each step, we use 

\begin{align*}
\Amat_{t+1} = \tfrac{1}{t+1} (t \Amat_t + \zvec_t \dvec_t^\top) =   (1-\beta_t) \Amat_t + \beta_t \zvec_t \dvec_t^\top
\end{align*}

\noindent
for $\beta_t = \frac{1}{t+1}$. 
The multiplication of $\Amat_t$ by $1-\beta_t$ requires only O($r$) computation
because $(1-\beta_t) \Umat_r \Sigmamat_r \Vmat_r^\top =  \Umat_r (1-\beta_t)\Sigmamat_r \Vmat_r^\top$.
Multiplying the full $\Amat$ matrix by $1-\beta_t$, on the other hand, would require
O$(\xdim^2)$ computation, which is prohibitive.
Further, $\beta_t$ can be selected to obtain a running average, as in Algorithm \ref{alg_tlstd},
or more generally can be set to any $\beta_t \in (0,1)$. 
For example, to improve tracking, $\beta$ can be chosen as a constant to weight more
recent samples more highly in the value function estimate. 

\begin{figure*}
\centering
	\subfigure[Rank versus performance with tile coding]{
	         \includegraphics[width=\figwidththree]{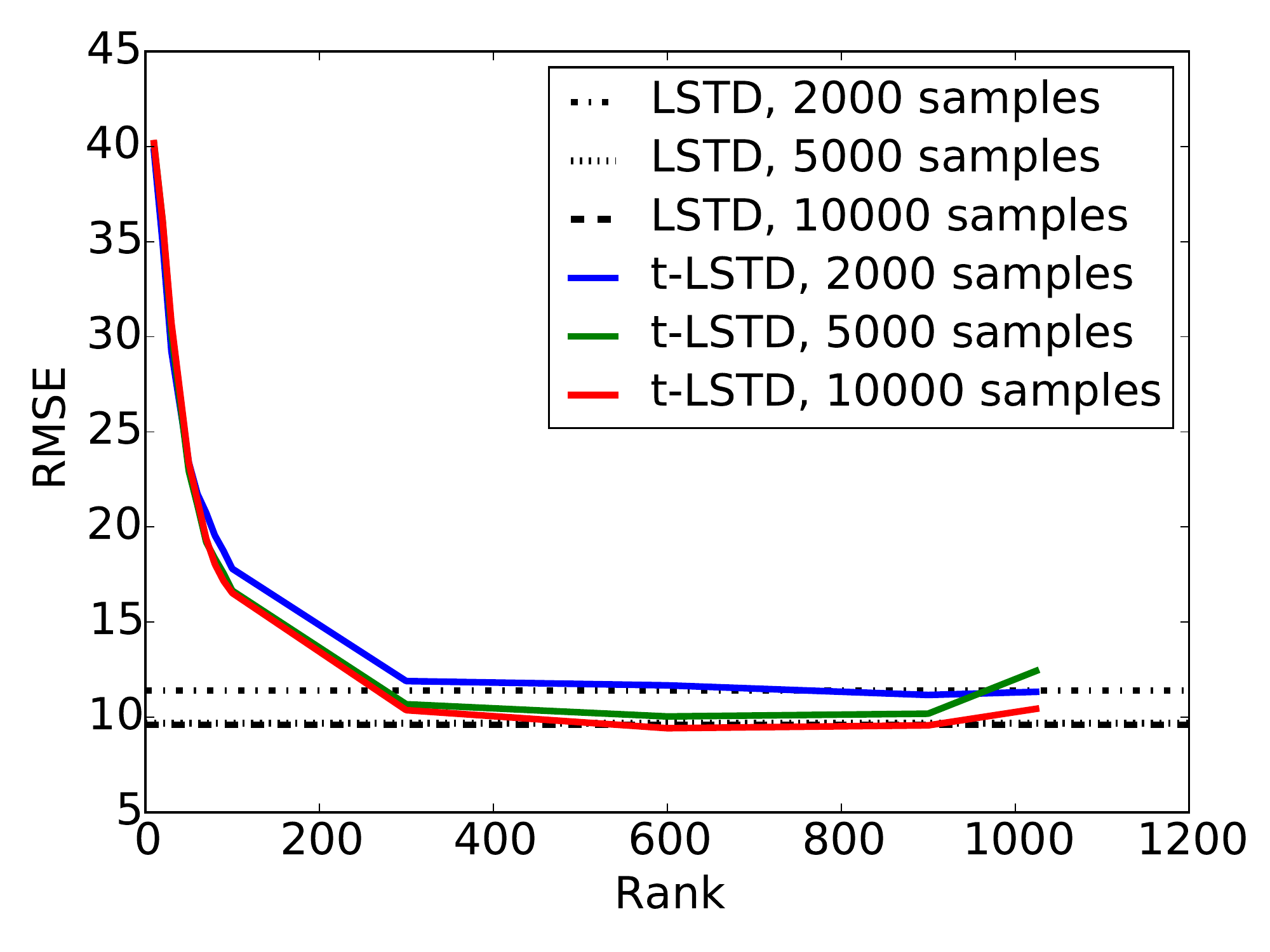}\label{fig:rank_tile}}
	\subfigure[Rank versus performance with RBFs]{
		\includegraphics[width=\figwidththree]{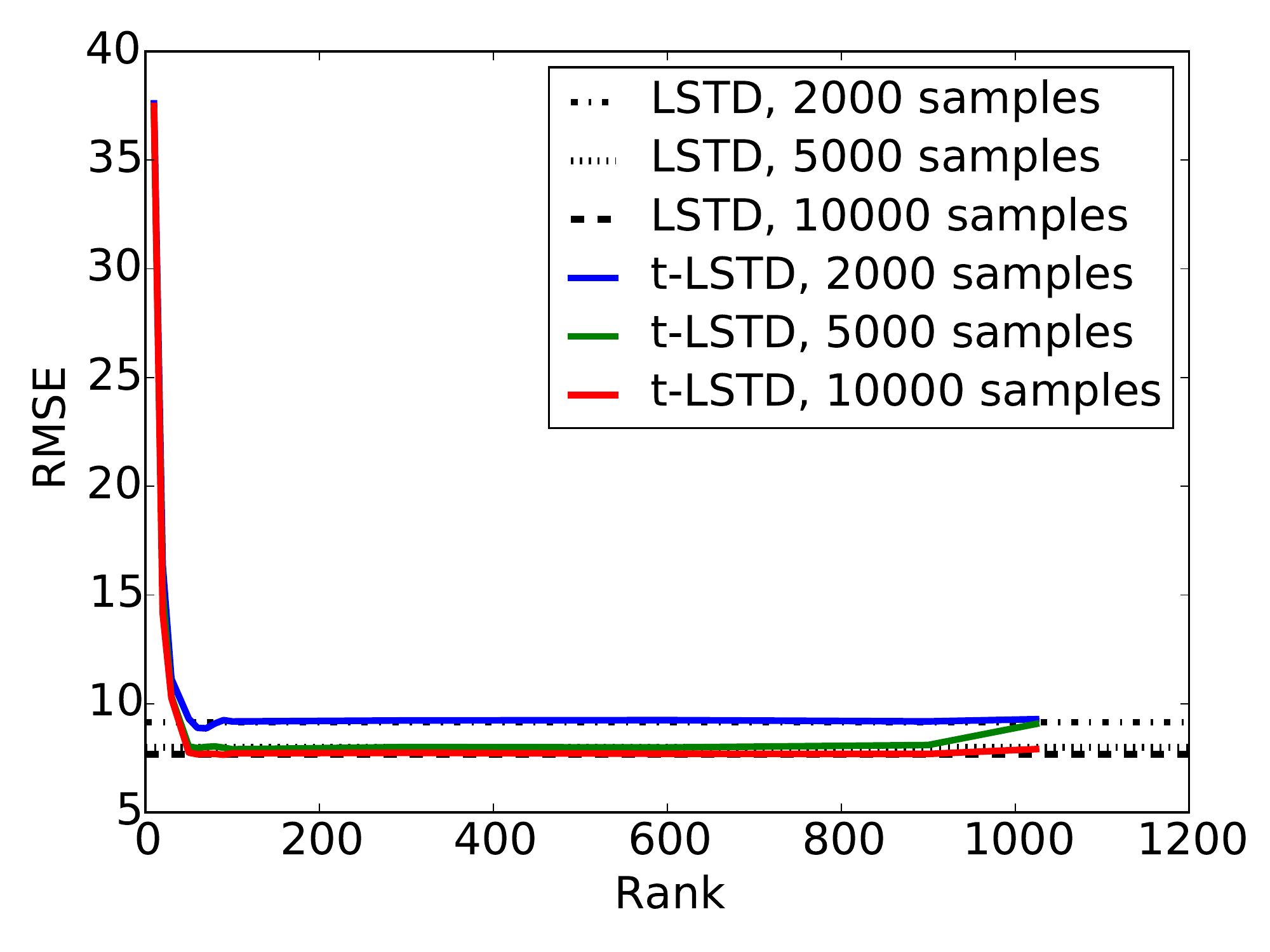}\label{fig:rank_rbf}}	
\begin{minipage}{\figwidththree}	
		\vspace{-4.0cm}	
	\caption{\small The impact of the rank $\rdim$ on RMSE of the true discounted returns and the learned value function in Mountain Car. 
	We can see that large $\rdim$ are not necessary, with performance levelling off at $\rdim = 50$. For high values of $\rdim$ and fewer samples, the error slightly increases, likely due to some instability with incremental updating and very small singular values. 
	}
	\label{fig:rmse-compare-rank}
	\end{minipage}\\
	\vspace{-0.2cm}
\subfigure[RMSE versus samples]{
		\includegraphics[width=\figwidththree]{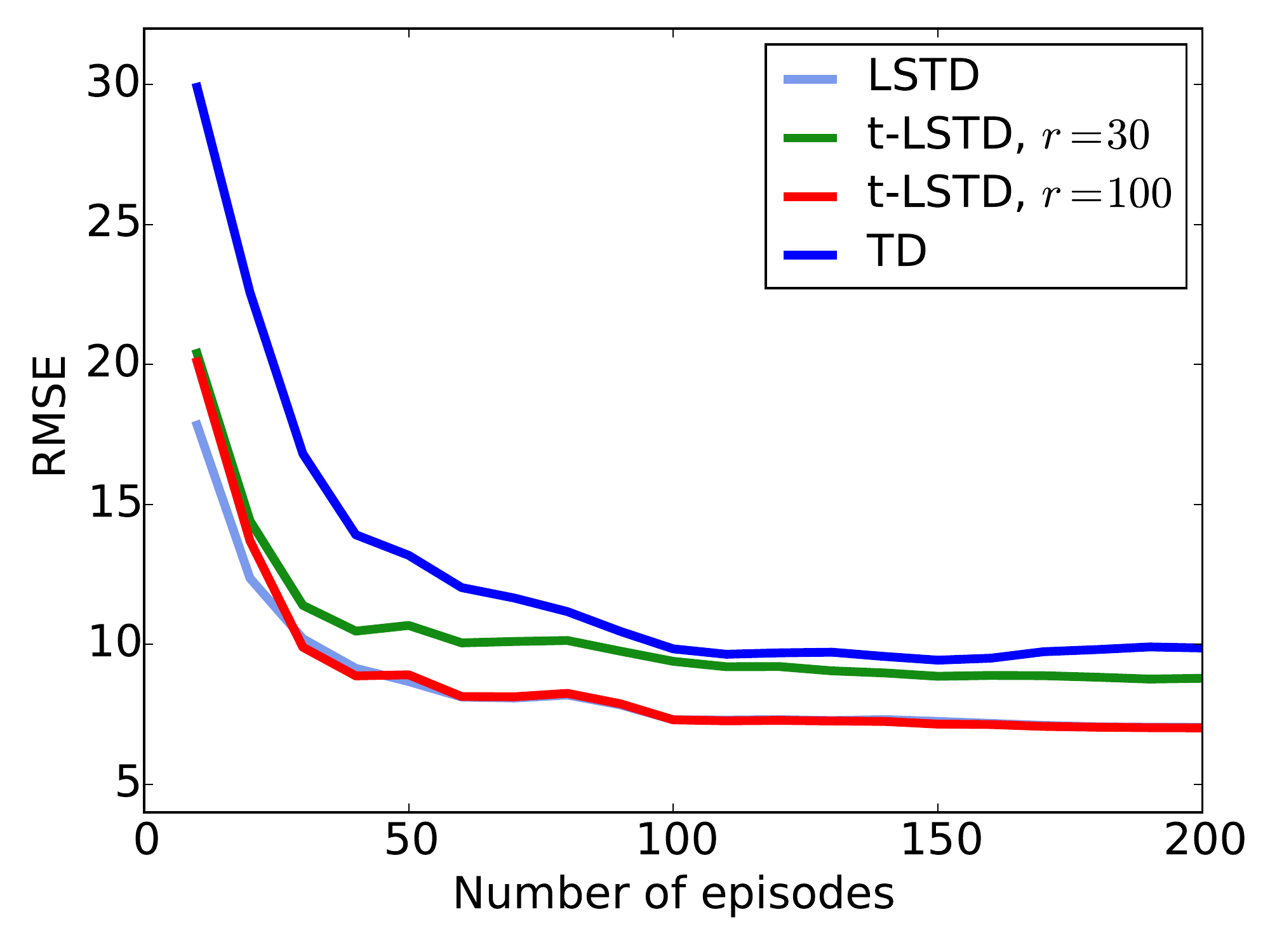}\label{fig:rbf-compare-mc}}			
\subfigure[RMSE versus runtime]{
		\includegraphics[width=\figwidththree]{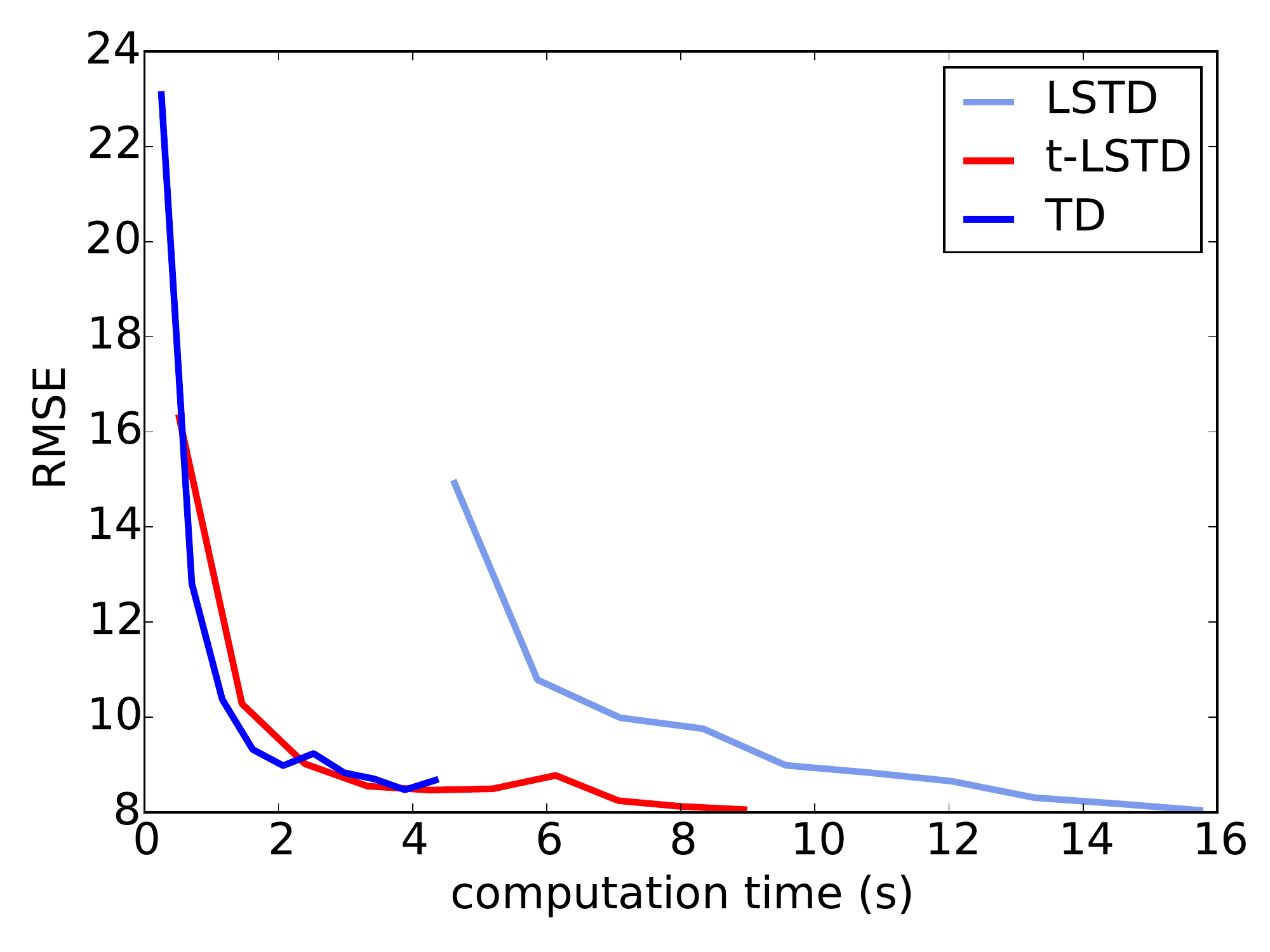}\label{fig:runtime}}	
\begin{minipage}{\figwidththree}	
		\vspace{-4.0cm}			
	\caption{\small RMSE of the true discounted returns and the learned value function
	 in Mountain Car with RBFs 
	\textbf{For (a)} we can see that with a significantly reduced $\rdim$, \tLSTD\ can
	match LSTD, and outperforms TD. This is the best setting for LSTD, where computation is
	not restricted, and it can spend time processing samples. 
	\textbf{For (b)} we provide the best scenario for TD, with unlimited samples.
	Once again, \tLSTD\ can
	almost match the performance of TD, and significantly outperforms LSTD. Together, these graphs indicate
	that \tLSTD\ can balance between the two extremes.
	The reported results are for the best parameter settings for TD, and for $r = 100$ and $\lambda = 0$ for \tLSTD.
	}
	\label{fig:rmse-compare-learning}
	\end{minipage}

\end{figure*}



\section{Experiments}
We empirically investigate
\tLSTD, for $k = \rdim$ in a benchmark domain 
and $k=1$ in an energy allocation domain. 

\vspace{0.1cm}
\noindent
\textbf{Value function accuracy in benchmark domains:}

We first investigate the performance of \tLSTD\ in the Mountain Car benchmark. 
The goal in this setting is to carefully investigate \tLSTD\
against the two extremes of TD and LSTD, and evaluate the utility
for balancing sample and computational complexity.
We use two common feature representations: tile coding and radial basis function (RBF) coding. We set the policy to the commonly used energy-pumping policy, which picks actions by pushing along the current velocity. The true values are estimated by using rollouts from states chosen in a uniform 20x20 grid of the state-space. The reported root mean squared error (RMSE) is computed between the estimated value functions and the rollout values.
The tile coding representation has 1000 features, using 10 layers of 10x10 grids.
The RBF representation has 1024 features, for a grid of 32x32 RBFs with width equal to $0.12$ times the total range of the state space. We purposefully set the total number of features to be similar in both cases in order to keep the results comparable.
We set the RBF widths to obtain good performance from LSTD. 
The other parameters ($\lambda$ and step-size) are optimized for each algorithm. 
In the Mountain Car results, we use the mini-batch case where $k=r$ and a discount $\gamma = 0.99$.
Results are averaged over 30 runs.

Empirically, we observed that the $\Amat$ has only a few large singular value with the rest being small. This was observed in Mountain Car across a wide range of parameter choices for both tile coding and RBFs, hinting that $\Amat$ could be reasonably approximated with small rank. In order to investigate the effect of the rank of \tLSTD\ , we vary $r$ and run \tLSTD\ on some fix number of samples. In Figure \ref{fig:rmse-compare-rank} (a) and (b), we observe a gracious decay in the quality of the estimated value function as the rank is reduced while achieving LSTD level performance with as little as $r=50$ for RBFs $(d=1024)$ and $r=300$ for tile coding $(d=1000)$.

Given large enough rank and numerical precision, LSTD and \tLSTD\ should behave similarly. To verify this, in Figure \ref{fig:rmse-compare-learning} (a), we plot the learning curves of \tLSTD\ in the case where the $r$ is too small and the case where $r$ is large enough, alongside LSTD and TD. As expected, we observe LSTD and \tLSTD\ to have near identical learning curves for $r=100$, while, for the case with smaller rank $r=30$, we see the algorithm converge rapidly to an inferior solution. TD is less sample efficient and so
converges more slowly than either.

Sample efficiency is an important property for an algorithm but does not completely capture the needs of an engineer attempting to solve a domain. In many case, the requirements tend to call for a balance between runtime and number of samples. In cases where a simulator is available, such as in game playing (e.g., atari, chess, backgammon, go), samples are readily available and only computational cost matters. For this reason, we explore the performance of TD, LSTD, and \tLSTD\ when given unlimited data but limited CPU time. 
In Figure \ref{fig:rmse-compare-learning} (b), we plot the accuracy of the methods with respect to computation time used. 
The algorithms are given access to varying amounts of samples: up to 8000 samples for TD and up to 4000 for \tLSTD\ and LSTD.
The RMSE and time taken is monitored, after which, the points are averaged to generate the plots comparing runtime to the error in the learned solution.


These results show that TD, despite poor sample efficiency, outperforms LSTD for a given runtime, due to the computational efficiency of each update. This supports the trend of preferring TD for large problems over LSTD. We observe \tLSTD\ achieve a comparable runtime to TD. Even though \tLSTD\ is computationally more costly than TD, its superior sample efficiency compensates.
Furthermore, this infinite sample stream case is favorable to TD. In a scenario where data is obtain in real-time, sacrificing sample efficiency for computational gains might leave TD idling occasionally, further reinforcing \tLSTD\ as a good alternative.


These results indicate that \tLSTD\ offers an effective approach to balance sample efficiency and computational efficiency to match both TD and LSTD in their respective use cases, offering good performance when data is plentiful while still offering LSTD-like sample efficiency.

\begin{figure*}[htp!]
	\centering
	\subfigure[RMSE vs samples, rank comparison]{
		\includegraphics[width=\figwidththree]{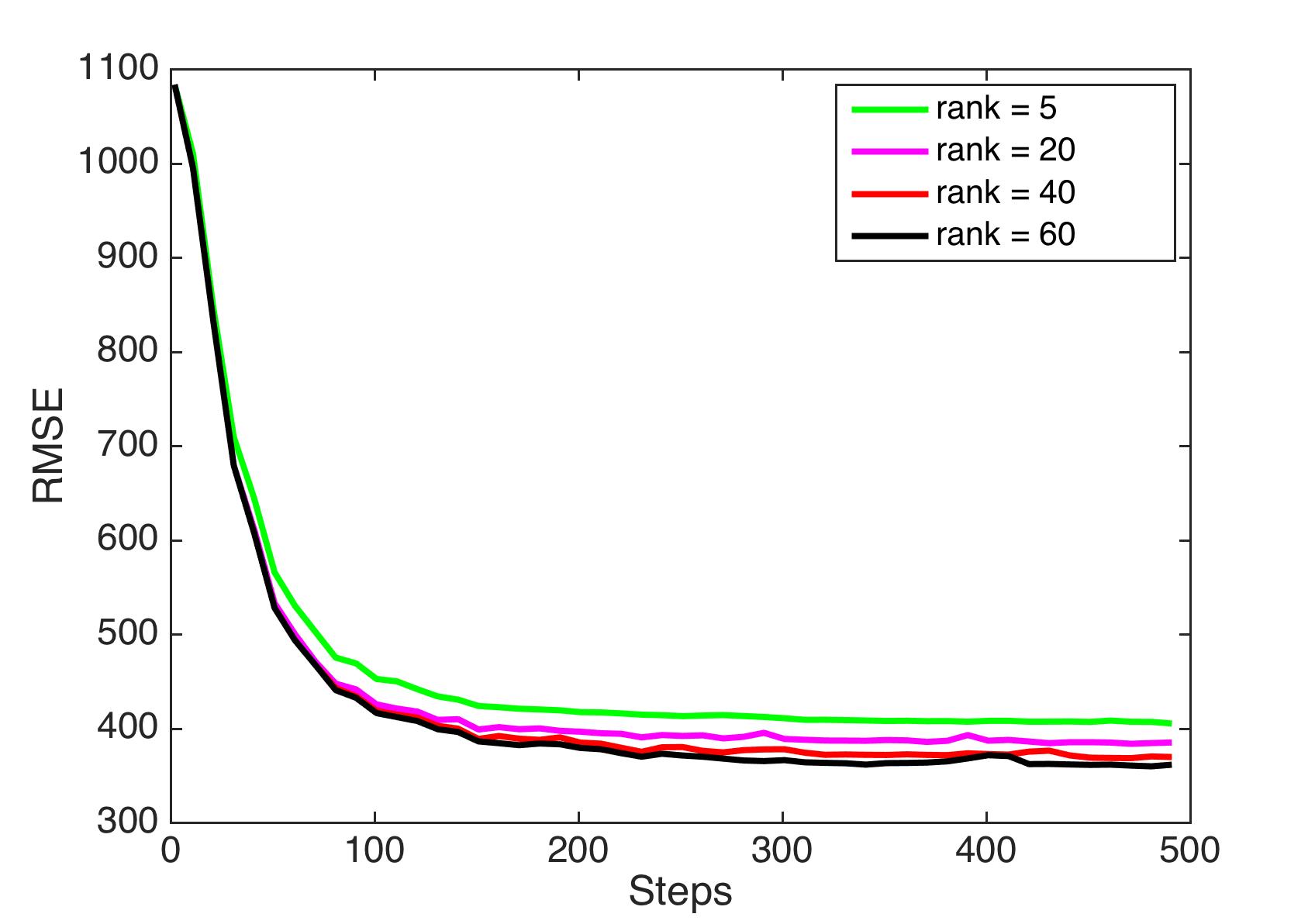}\label{fig:tlstd_r}}
	\subfigure[RMSE vs samples, algorithm comparison]{
		\includegraphics[width=\figwidththree]{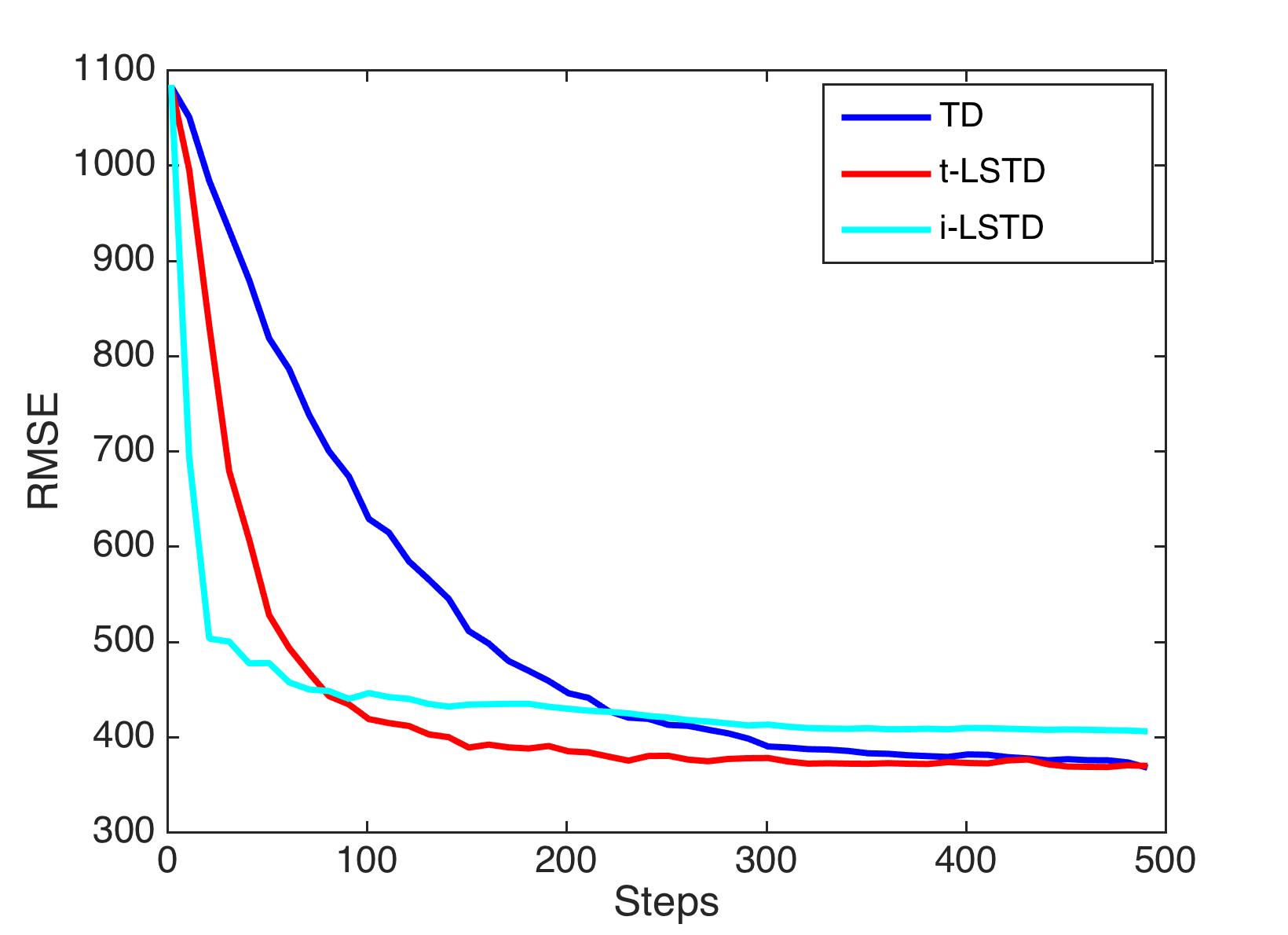} \label{fig:compare_three_algo}}
	\subfigure[RMSE vs runtime]{
		\includegraphics[width=\figwidththree]{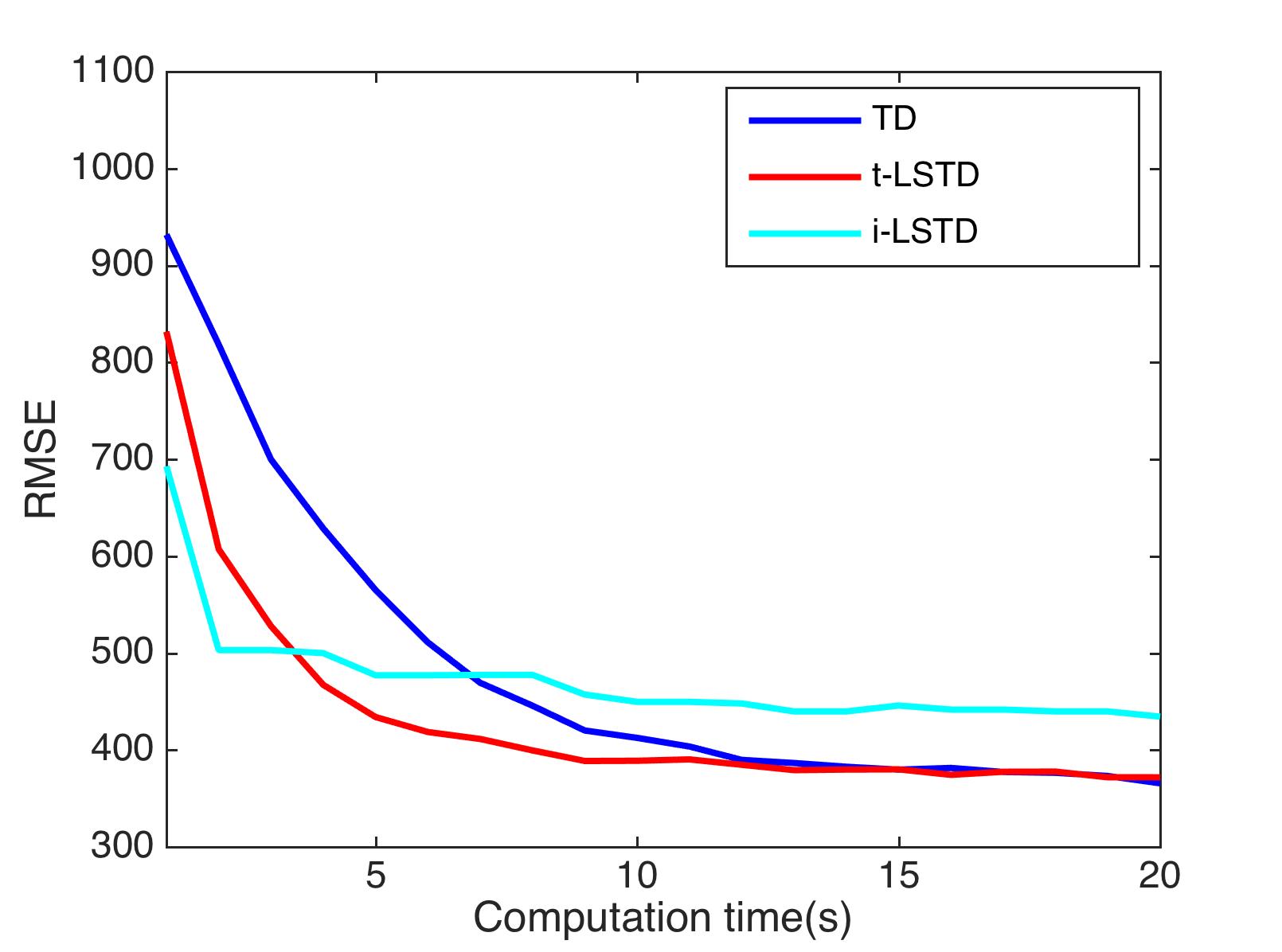}\label{fig:compare_three_algo_runtime}}
   	\caption{ 
RMSE of the value function in the energy allocation domain. 
\textbf{(a)} 
Performance of \tLSTD\ improves as the rank increases; however, even for small $\rdim = 5$, the algorithm still converges with some bias. 
For $\rdim$ smaller than $5$, the error was significantly worse. 
\textbf{(b)} With $\rdim = 40$,
\tLSTD\ converges to the almost same level with TD in significantly fewer steps. The best parameters are chosen for each algorithm, m = 50 for iLSTD and r = 40 for tLSTD. 
\textbf{(c)} As before, we plot RMSE versus runtime, but now by selecting a
scenario in between the extremes plotted in Figure \ref{fig:rmse-compare-learning}.
The number of samples per second is restricted to 25 samples, meaning TD is
sometimes idle waiting for more samples, and iLSTD (m = 50) and \tLSTD\ (r = 40) could be too slow to process all the samples. This plot further indicates the advantages of \tLSTD,
particularly as it is faster than TD in terms of sample efficiency and scales better than iLSTD and converges to a better solution.  
}\label{fig_energy}
\end{figure*}


\vspace{0.1cm}
\noindent
\textbf{Value function accuracy in an energy domain}


In this section, we demonstrate the performance of the fully incremental algorithm ($\updelay = 1$) in a large energy allocation domain \citep{salas2013benchmarking}. 
The focus in this experiment is to evaluate the practical
utility of \tLSTD\ in an important application, versus realistic competitors:
TD\footnote{We also compared to true-online TD \cite{vanseijen2014true}, but it gave very similar performance; we therefore omit it.} and iLSTD. 
The goal of the agent in this domain is to maximize revenue and satisfy demand.
Each action vector is an allocation decision. 
Each state is a four dimensional variable: the amount of energy in storage, the amounts of renewable generation available, the market price of energy, and the demand needs to be satisfied.
We use a provided near-optimal policy \citep{salas2013benchmarking}.
We set $\gamma = 0.8$.

To approximate the value function, we use tile coding with 
$32$ tilings where each tiling contains $5 \times 5 \times 10 \times 5$ grids,
resulting in $40,000$ features and also included a bias unit. 
We choose this representation, because iLSTD is only computationally feasible 
for high-dimensional \textit{sparse} representations.
As before, extensive rollouts are computed from a subset of states, to compute
accurate estimate of the true value, and then stored for comparison
in the computation of the RMSE. Results were averaged over 30 runs.

We report results for several values of $\rdim$ for \tLSTD.
We sweep the additional parameters in the other algorithms,
including step-sizes $\alpha$ for TD and iLSTD and $m$ for iLSTD. 
We sweep a range of $\alpha_0 = \{2^{-11}, 2^{-10}, 2^{-9}, ..., 2^{-1}\}$,
and divide by the number of active features (which in this case is $2^6$).
Further, because iLSTD is unstable unless $\alpha$ is decayed,
we further sweep the decay formula as suggested by \citet{geramifard2006incremental}
\ralignspace
\begin{align*}
\alpha_t = \alpha_0 \frac{N_0 + 1}{N_0 + t},
\end{align*}
\balignspace
 \noindent
 where $N_0$  is chosen from $\{10, 100, 1000\}$.
 To focus parameter sweeps on the step-size, which had much more effect
 for iLSTD, we set $\lambda = 0.9$ for all other algorithms, except for tLSTD which we set $\lambda = 1.0$.
We choose $\rdim \in \{5, 20, 40, 60\}$ and $m \in \{10,20,30,40,50\}$. We restrict the iLSTD parameters to a small set, since there are too many options, even the optimal stepsize would be different when we choose different m values. Preliminary investigation indicated that $50$ was large enough for iLSTD.

In this domain, under the common strategy of creating a large number
of fixed features (tile coding or RBFs), \tLSTD\ is able to significantly take
advantage of low rank structure, learning more efficiently without 
incurring much computational cost. 
Figure \ref{fig_energy} shows that \tLSTD\ performs
well with a small $\rdim = 40 << \xdim = 40,000$,  
and outperforms both TD and iLSTD. 


We highlight that iLSTD is one of the only
practical competitors introduced for this setting:
incremental learning with computational constraints.
Even then, iLSTD is restrictive in that the feature representation
must be sparse and its storage requirements are O($\xdim^2$). 
Further, though it was reasonably robust to the choice of $m$,
we found iLSTD was quite sensitive to the choice of step-size parameter.
In fact, without a careful decay, we still encountered divergence issues.

The goal here was to investigate the performance of the simplest version of \tLSTD, 
with fewest parameters and without optimizing thresholds, which were kept fixed
at reasonable heuristics across all experiments. This choice does impact the learning curves of \tLSTD. 
For example, though \tLSTD\ has significantly faster early convergence,
it is less smooth than either TD or iLSTD. 
This lack of smoothness could be due to not optimizing these parameters and further 
because $\wvec_t$ is solved on each step.
Beyond this vanilla implementation of
\tLSTD, there are clear avenues to explore to more smoothly update $\wvec_t$ with the low-rank approximation to $\Amat$.
Nonetheless, even in its simplest form, \tLSTD\ provides an attractive alternative to TD,
obtaining sample efficiency improvements without much additional computation
and without the need to tune a step-size parameter. 

\section{Discussion and conclusion}

This paper introduced an efficient value function approximation algorithm, called \tLSTD($\lambda$), that maintains
an incremental truncated singular value decomposition of the LSTD matrix. 
We systematically explored the validity of using low-rank approximations
for LSTD, first by proving a simulation error bound for truncated low-rank LSTD solutions
and then, empirically, by examining an incremental truncated LSTD algorithm in two domains.
We demonstrated performance of \tLSTD\ in the benchmark domain, Mountain Car, exploring runtime properties and the effect of the small rank approximation,
and in a high-dimensional energy allocation domain, 
illustrating that \tLSTD\ enables a nice interpolation between the properties
of TD and LSTD, and out-performs iLSTD. 

There are several potential benefits of \tLSTD\ that we did not
yet explore in this preliminary investigation. 
First, there are clear advantages to \tLSTD\ for
tracking and control. As mentioned above, unlike previous LSTD algorithms, 
past samples for \tLSTD\ can be efficiently down-weighted with a $\beta_t \in (0,1)$. 
By enabling down-weighting, $\Amat_t$ is more strongly
influenced by recent samples and so can better adapt in
a non-stationary environment, such as for control. 

Another interesting avenue is to take advantage of \tLSTD\
for early learning, to improve sample efficiency,
and then switch to TD to converge to an unbiased solution. 
Even for highly constrained systems in terms of storage and computation,
aggressively small $\rdim$ can still be useful for early learning.
Further empirical investigation
could give insight into when this switch could occur, depending on the choice of $\rdim$.

Finally, an important avenue for this new approach 
is to investigate
the convergence properties
of truncated incremental SVDs. 
The algorithm derivation requires only simple algebra and is clearly sound;
however, to the best of our knowledge, the question of convergence under numerical stability 
 and truncating non-zero singular values 
 remains open.
 The truncated incremental SVD has been
 shown to be practically useful in numerous occasions, 
 such as for principal components analysis and partial least squares \citep{arora2012stochastic}.
 Moreover, there are some informal arguments (using randomized matrix theory)
 that even under truncation the SVD will re-orient \citep{brand2006fast}. 
 This open question is 
 an important next step in understanding $\tLSTD$ and, more generally, incremental singular value decomposition algorithms
 for reinforcement learning.



{
\bibliographystyle{named}
\bibliography{paper}

\begin{thebibliography}{}

\bibitem[\protect\citeauthoryear{Arora \bgroup \em et al.\egroup
  }{2012}]{arora2012stochastic}
R~Arora, A~Cotter, K~Livescu, and N~Srebro.
\newblock {Stochastic optimization for PCA and PLS}.
\newblock In {\em Annual Allerton Conference on Communication, Control, and
  Computing}, 2012.

\bibitem[\protect\citeauthoryear{Bertsekas}{2007}]{bertsekas2007dynamic}
D~Bertsekas.
\newblock {\em {Dynamic Programming and Optimal Control}}.
\newblock Athena Scientific Press, 2007.

\bibitem[\protect\citeauthoryear{Bowling and
  Geramifard}{2008}]{bowling2008sigma}
M~Bowling and A~Geramifard.
\newblock {Sigma point policy iteration}.
\newblock In {\em International Conf. on Autonomous Agents and Multiagent
  Systems}, 2008.

\bibitem[\protect\citeauthoryear{Boyan}{1999}]{boyan1999least}
J~A Boyan.
\newblock {Least-squares temporal difference learning}.
\newblock {\em International Conf. on Machine Learning}, 1999.

\bibitem[\protect\citeauthoryear{Bradtke and Barto}{1996}]{bradtke1996linear}
Steven~J Bradtke and Andrew~G Barto.
\newblock {Linear least-squares algorithms for temporal difference learning}.
\newblock {\em Machine Learning}, 1996.

\bibitem[\protect\citeauthoryear{Brand}{2006}]{brand2006fast}
Matthew Brand.
\newblock {Fast low-rank modifications of the thin singular value
  decomposition}.
\newblock {\em Linear Algebra and its Applications}, 2006.

\bibitem[\protect\citeauthoryear{Eckart and Young}{1936}]{eckhart1936the}
Carl Eckart and Gale Young.
\newblock {The approximation of one matrix by another of lower rank}.
\newblock {\em Psychometrika}, 1936.

\bibitem[\protect\citeauthoryear{Farahmand \bgroup \em et al.\egroup
  }{2008}]{farahmand2008regularized}
A~M Farahmand, M~Ghavamzadeh, and C~Szepesv{\'a}ri.
\newblock {Regularized policy iteration}.
\newblock In {\em Advances in Neural Information Processing Systems}, 2008.

\bibitem[\protect\citeauthoryear{Farahmand}{2011}]{farahmand2011thesis}
A~Farahmand.
\newblock {\em {Regularization in reinforcement learning}}.
\newblock PhD thesis, Univ. of Alberta, 2011.

\bibitem[\protect\citeauthoryear{Fong and Saunders}{2011}]{fong2011lsmr}
David Chin-Lung Fong and Michael Saunders.
\newblock {LSMR: An Iterative Algorithm for Sparse Least-Squares Problems}.
\newblock {\em SIAM Journal on Scientific Computing}, 2011.

\bibitem[\protect\citeauthoryear{Geramifard and
  Bowling}{2006}]{geramifard2006incremental}
A~Geramifard and M~Bowling.
\newblock {Incremental least-squares temporal difference learning}.
\newblock In {\em AAAI Conference on Artificial Intelligence}, 2006.

\bibitem[\protect\citeauthoryear{Geramifard \bgroup \em et al.\egroup
  }{2007}]{geramifard2007ilstd}
A~Geramifard, M~Bowling, and M~Zinkevich.
\newblock {iLSTD: Eligibility traces and convergence analysis}.
\newblock In {\em Advances in Neural Information Processing Systems}, 2007.

\bibitem[\protect\citeauthoryear{Ghavamzadeh and
  Lazaric}{2011}]{ghavamzadeh2011finite}
M~Ghavamzadeh and A~Lazaric.
\newblock {Finite-sample analysis of Lasso-TD}.
\newblock In {\em International Conference on Machine Learning}, 2011.

\bibitem[\protect\citeauthoryear{Ghavamzadeh \bgroup \em et al.\egroup
  }{2010}]{ghavamzadeh2010lstd}
M~Ghavamzadeh, A~Lazaric, O~A Maillard, and R~Munos.
\newblock {LSTD with random projections}.
\newblock In {\em Advances in Neural Information Processing Systems}, 2010.

\bibitem[\protect\citeauthoryear{Golub and Kahan}{1965}]{golub1965calculating}
G~Golub and W~Kahan.
\newblock {Calculating the Singular Values and Pseudo-Inverse of a Matrix}.
\newblock {\em Journal of the Society for Industrial and Applied Mathematics
  Series B Numerical Analysis}, 1965.

\bibitem[\protect\citeauthoryear{Groetsch}{1984}]{groetsch1984thetheory}
C~W Groetsch.
\newblock {\em {The Theory of Tikhonov Regularization for Fredholm Equations of
  the First Kind}}.
\newblock Pitman Advanced Publishing Program, 1984.

\bibitem[\protect\citeauthoryear{Hansen}{1986}]{hansen1986thetruncated}
P~C Hansen.
\newblock {The truncated SVD as a method for regularization}.
\newblock {\em BIT Numerical Mathematics}, 1986.

\bibitem[\protect\citeauthoryear{Hansen}{1990}]{hansen1990thediscrete}
Per~Christian Hansen.
\newblock {The discrete picard condition for discrete ill-posed problems}.
\newblock {\em BIT Numerical Mathematics}, 1990.

\bibitem[\protect\citeauthoryear{Keller \bgroup \em et al.\egroup
  }{2006}]{keller2006automatic}
PW~Keller, S~Mannor, and D~Precup.
\newblock {Automatic basis function construction for approximate dynamic
  programming and reinforcement learning}.
\newblock In {\em International Conference on Machine Learning}, 2006.

\bibitem[\protect\citeauthoryear{Kolter and
  Ng}{2009}]{kolter2009regularization}
JZ~Kolter and AY~Ng.
\newblock {Regularization and feature selection in least-squares temporal
  difference learning}.
\newblock In {\em Inter. Conf. on Machine Learning}, 2009.

\bibitem[\protect\citeauthoryear{Lazaric \bgroup \em et al.\egroup
  }{2010}]{lazaric2010finite}
A~Lazaric, M~Ghavamzadeh, and R~Munos.
\newblock {Finite sample analysis of LSTD}.
\newblock {\em International Conference on Machine Learning}, 2010.

\bibitem[\protect\citeauthoryear{Lin}{1993}]{lin1993reinforcement}
Long-Ji Lin.
\newblock {\em {Reinforcement Learning for Robots Using Neural Networks}}.
\newblock PhD thesis, CMU, 1993.

\bibitem[\protect\citeauthoryear{Mirsky}{1960}]{mirsky1960symmetric}
L~Mirsky.
\newblock {Symmetric gauge functions and unitarily invariant norms}.
\newblock {\em Quartely Journal Of Mathetmatics}, 1960.

\bibitem[\protect\citeauthoryear{Pires and
  Szepesvari}{2012}]{pires2012statistical}
Bernardo~Avila Pires and Csaba Szepesvari.
\newblock {Statistical linear estimation with penalized estimators: an
  application to reinforcement learning}.
\newblock In {\em International Conference on Machine Learning}, 2012.

\bibitem[\protect\citeauthoryear{Prashanth \bgroup \em et al.\egroup
  }{2013}]{prashanth2013fast}
L~A Prashanth, Nathaniel Korda, and R{\'e}mi Munos.
\newblock {Fast LSTD using stochastic approximation: Finite time analysis and
  application to traffic control}.
\newblock {\em arXiv.org}, 2013.

\bibitem[\protect\citeauthoryear{Roman \bgroup \em et al.\egroup
  }{2008}]{roman2008arobust}
J~E Roman, Vicente Hernandez, and Andres Tomas.
\newblock {A robust and efficient parallel SVD solver based on restarted
  Lanczos bidiagonalization}.
\newblock {\em Electronic Transactions on Numerical Analysis}, 2008.

\bibitem[\protect\citeauthoryear{Salas and
  Powell}{2013}]{salas2013benchmarking}
D~F Salas and W~B Powell.
\newblock {Benchmarking a Scalable Approximate Dynamic Programming Algorithm
  for Stochastic Control of Multidimensional Energy Storage Problems}.
\newblock {\em Dept Oper Res Financial Eng}, 2013.

\bibitem[\protect\citeauthoryear{Sutton and
  Barto}{1998}]{sutton1998reinforcement}
R.S. Sutton and A~G Barto.
\newblock {\em {Reinforcement Learning: An Introduction}}.
\newblock MIT press, 1998.

\bibitem[\protect\citeauthoryear{Sutton}{1988}]{sutton1988learning}
R.S. Sutton.
\newblock {Learning to predict by the methods of temporal differences}.
\newblock {\em Machine Learning}, 1988.

\bibitem[\protect\citeauthoryear{Tagorti and
  Scherrer}{2015}]{tagorti2015ontherate}
Manel Tagorti and Bruno Scherrer.
\newblock {On the Rate of Convergence and Error Bounds for LSTD(\(\lambda\)).}
\newblock In {\em Inter. Conf. on Machine Learning}, 2015.

\bibitem[\protect\citeauthoryear{Tsitsiklis and
  Van~Roy}{1997}]{tsitsiklis1997ananalysis}
J.N. Tsitsiklis and B~Van~Roy.
\newblock {An analysis of temporal-difference learning with function
  approximation}.
\newblock {\em IEEE Transactions on Automatic Control}, 1997.

\bibitem[\protect\citeauthoryear{van Seijen and
  Sutton}{2014}]{vanseijen2014true}
Harm van Seijen and Rich Sutton.
\newblock {True online TD(lambda)}.
\newblock In {\em International Conference on Machine Learning}, 2014.

\bibitem[\protect\citeauthoryear{van Seijen and
  Sutton}{2015}]{vanseijen2015adeeper}
H~van Seijen and R.S. Sutton.
\newblock {A deeper look at planning as learning from replay}.
\newblock In {\em International Conference on Machine Learning}, 2015.

\end{thebibliography}
}

\newpage
\onecolumn
\appendix

\section{Proof of Theorem \ref{thm_tradeoff}}

\textbf{Theorem \ref{thm_tradeoff}}[Bias-variance trade-off of rank-$\rdim$ approximation]
Let $\Amat_{t,\rdim} =  \hat{\Umat} \hat{\Sigmamat}_{\rdim} \hat{\Vmat}^\top$ be the approximated $\Amat$
after $t$ samples, truncated to rank $\rdim$, i.e., with the last $\rdim+1, \ldots, \xdim$ singular values zeroed.
Let $\wvec^* = \Amat^\pinv \bvec$ and $\wvec_{t,\rdim} = \Amat_{t,\rdim}^\pinv \bvec_t$.
Under Assumption 1 and 2, the relative error of the rank-$\rdim$ weights to the true weights $\wvec^*$ is bounded as follows:
\begin{align*}
\| \wvec_{t,\rdim} - \wvec^* \|_2
  &\le
   \frac{1}{\sighat{t}{\rdim}} \| \bvec_t - \Amat_t \wvec^*  \|_2 
   + 
 (\xdim - \rdim)  \epsilon(t)
    +
 \underbrace{ (\xdim - \rdim) \sigma_{\rdim}^{\picardpower - 1}}_{\text{bias}}
\end{align*}
for a function $\epsilon: \NN \rightarrow [0, \rankA\sigma_{1}^{\picardpower-1}]$, where
$\epsilon(t) \rightarrow 0$ as $t \rightarrow \infty$:
\begin{align*}
\epsilon(t) = \min \left(
   \rankA  \sigma_{1}^{\picardpower-1}, \sum_{j=1}^\rankA \sqnorm{\vvec_j \sigma_j^{p-1} - \hat{\vvec}_j \sighat_j^{p-1}} + \sighat_\rdim^{p-1} - \sigma_\rdim^{p-1}\right)
.
\end{align*}
\begin{proof}
We split up the error into two terms: 
approximation error due to a finite number of samples $t$
and
bias due the choice of $\rdim < \xdim$.
Let $\dvec_t = \bvec_t - \Amat_t \wvec^*$ and notice that 
\ralignspace
\begin{align*}
\Amat_{t,\rdim}^\pinv \Amat_{t} 
&= \hat{\Vmat} \hat{\Sigmamat}^\inv_{\rdim} \hat{\Umat}^\top\hat{\Umat} \hat{\Sigmamat} \hat{\Vmat}^\top
= \hat{\Vmat} \hat{\Sigmamat}^\inv_{\rdim}  \hat{\Sigmamat} \hat{\Vmat}^\top
= \sum_{i=1}^\rdim \hat{\vvec}_i \hat{\vvec}_i^\top
.
\end{align*}
\reducealignspace{-0.3cm}
\noindent
Therefore, because $\sum_{i=1}^\xdim \hat{\vvec}_i \hat{\vvec}_i^\top = \eye$
\begin{align*}
 \wvec_{t,\rdim} - \wvec^* 
 &= \Amat_{t,\rdim}^\pinv \bvec_t - \wvec^*\\
& = \Amat_{t,\rdim}^\pinv (\bvec_t - \Amat_t \wvec^*)+ \Amat_{t,\rdim}^\pinv \Amat_{t} \wvec^*  - \wvec^* 
 \\
 &= \Amat_{t,\rdim}^\pinv \dvec_t +  \sum_{i=1}^\rdim \hat{\vvec}_i \hat{\vvec}_i^\top \wvec^*  - \sum_{i=1}^\xdim \hat{\vvec}_i \hat{\vvec}_i^\top\wvec^*\\
 &= \Amat_{t,\rdim}^\pinv \dvec_t - \sum_{i=\rdim+1}^\xdim \hat{\vvec}_i \hat{\vvec}_i^\top\wvec^*
\end{align*}
Taking the $\ell_2$ norm of both sides, we get
\begin{align*}
 \| \wvec_{t,\rdim} - \wvec^* \|_2
  &\le
     \| \Amat_{t,\rdim}^\pinv\|_2 \| \dvec_t  \|_2 + \sqnorm{\sum_{i=\rdim+1}^\xdim \hat{\vvec}_i \hat{\vvec}_i^\top\wvec^*}\\
     &=
   \frac{1}{\sighat{t}{\rdim}} \| \dvec_t  \|_2 + \sqnorm{ \sum_{i=\rdim+1}^\xdim \hat{\vvec}_i \hat{\vvec}_i^\top\wvec^*}
\end{align*}
where the induced 2-norm on a matrix is the spectral norm (the largest singular value).

Now we can simplify the second term using 
$\wvec^* = \Vmat \Sigma^{\pinv}  \Umat^\top \bvec 
= \sum_{j=1}^\rankA \sigma_j^\inv  \vvec_j \uvec_j^\top \bvec$
and the fact that 
$\hat{\vvec}_i$ are orthonormal vectors, giving
$\| \sum_i \hat{\vvec}_i s_i \|_2^2 = s_1\hat{\vvec}_1^\top\hat{\vvec}_1s_1 + \ldots = \sum_i \|s_i \|_2$ for scalars $s_i$, 
%
\begin{align*}
\sqnorm{ \sum_{i=\rdim+1}^\xdim \hat{\vvec}_i \hat{\vvec}_i^\top\wvec^* }^2
&= \sqnorm{\sum_{i=\rdim+1}^\xdim \hat{\vvec}_i  \sum_{j=1}^\rankA \hat{\vvec}_i^\top \vvec_j \sigma_j^\inv   \uvec_j^\top \bvec}^2\\
&= \sum_{i=\rdim+1}^\xdim\sqnorm{ \sum_{j=1}^\rankA \hat{\vvec}_i^\top \vvec_j \sigma_j^\inv   \uvec_j^\top \bvec}^2
.
\end{align*}
Because $\| \xvec \|_2 = \sqrt{\sum x_i^2} \le \| \xvec \|_1 = \sum | x_i|$, we can apply the square root to each term
in the sum
\begin{align*}
\sqnorm{ \sum_{i=\rdim+1}^\xdim \hat{\vvec}_i \hat{\vvec}_i^\top\wvec^* }
&\le \sum_{i=\rdim+1}^\xdim\sqnorm{ \sum_{j=1}^\rankA \hat{\vvec}_i^\top \vvec_j \sigma_j^\inv   \uvec_j^\top \bvec}
.
\end{align*}
We will get two upper bounds, and take the minimum to obtain a tighter upper bound for early samples.
For the first upper bound
\begin{align*}
\sum_{i=\rdim+1}^\xdim\sqnorm{ \sum_{j=1}^\rankA \hat{\vvec}_i^\top \vvec_j \sigma_j^\inv   \uvec_j^\top \bvec}
&= \sum_{i=\rdim+1}^\xdim \sqnorm{ \hat{\vvec}_i^\top \vvec_i \sigma_i^\inv   \uvec_i^\top \bvec
 +  \sum_{j=1, j \neq i}^\rankA \hat{\vvec}_i^\top \vvec_j \sigma_j^\inv   \uvec_j^\top \bvec} \\
  &\le  \sum_{i=\rdim+1}^\xdim \!\!\|\hat{\vvec}_i^\top\|_2  \|\vvec_i\|_2 \sigma_i^{\picardpower - 1}
 \!\!+ \!\sum_{i=\rdim+1}^\xdim\sum_{j=1, j \neq i}^\rankA \!\!\! \|\hat{\vvec}_i^\top \vvec_j\|_2  \sigma_j^{\picardpower-1}\\
   &\le   (\xdim - \rdim) \sigma_\rdim^{\picardpower - 1}
 + (\xdim - \rdim) \tilde{\epsilon}(t) \rankA  \sigma_{1}^{\picardpower-1}
\end{align*}
where $\tilde{\epsilon}(t)$ is some function with $\|\hat{\vvec}_i^\top \vvec_j\|_2 \le \tilde{\epsilon}(t) \le 1$. 
The first inequality follows from using the triangle inequality and the discrete Picard condition.
Further, we know that $\tilde{\epsilon}(t) \le 1$ due to the fact that $\|\hat{\vvec}_i^\top \vvec_j\|_2  \le 1$
because they are both unit vectors. We will further quantity the second term including $\tilde{\epsilon}(t)$
below, to better understand how quickly this term disappears. 

In general, even without an explicit rate of convergence, 
we know that there exists a
function $\tilde{\epsilon}: \NN \rightarrow [0,1]$ such that $\tilde{\epsilon}(t) \rightarrow 0$ as $t \rightarrow \infty$
and
$| \hat{\vvec}_i^\top \vvec_j | \le \tilde{\epsilon}(t) \ \ \ \forall j = 1, \ldots, \rank(\Amat), i = 1 \ldots, \xdim \text{ with } i \neq j$
.
This follows because $\Amat_t \rightarrow \Amat$ 
and so we know that $\hat{\vvec}_j \rightarrow \vvec_j$, for the singular vectors
that correspond to non-zero singular values, $j \le \rank(\Amat)$. 
The other singular vectors 
must be orthogonal to the $\hat{\vvec}_j$ for $j \le \rank(\Amat)$. 
Consequently, 
$| \hat{\vvec}_i^\top \vvec_j | \rightarrow 0$ as $t \rightarrow \infty$ for $j \le \rank(\Amat), i\neq j$.

For the second upper bound, 
we further quantify $\epsilon(t)$.
%
\begin{align*}
\sum_{i=\rdim+1}^\xdim\sqnorm{ \sum_{j=1}^\rankA \hat{\vvec}_i^\top \vvec_j \sigma_j^\inv   \uvec_j^\top \bvec}
&\le \sum_{i=\rdim+1}^\xdim \sum_{j=1}^\rankA \sqnorm{\hat{\vvec}_i^\top \vvec_j \sigma_j^\inv   \uvec_j^\top \bvec}\\
&\le \sum_{i=\rdim+1}^\xdim \sum_{j=1}^\rankA \sqnorm{\hat{\vvec}_i^\top \vvec_j \sigma_j^\inv}   \sigma_j^p\\
&= \sum_{i=\rdim+1}^\xdim \sum_{j=1}^\rankA \sqnorm{\hat{\vvec}_i^\top \vvec_j \sigma_j^{p-1}} \\
&= \sum_{i=\rdim+1}^\xdim \sum_{j=1}^\rankA \sqnorm{\hat{\vvec}_i^\top (\vvec_j \sigma_j^{p-1} - \hat{\vvec}_j \sighat_j^{p-1}) + \hat{\vvec}_i^\top \hat{\vvec}_j \sighat_j^{p-1}} \\
&\le \sum_{i=\rdim+1}^\xdim \sum_{j=1}^\rankA \sqnorm{\hat{\vvec}_i^\top (\vvec_j \sigma_j^{p-1} - \hat{\vvec}_j \sighat_j^{p-1})} + \sum_{i=\rdim+1}^\xdim \sum_{j=1}^\rankA\sqnorm{\hat{\vvec}_i^\top \hat{\vvec}_j \sighat_j^{p-1}} \\
&= \sum_{i=\rdim+1}^\xdim \sum_{j=1}^\rankA \sqnorm{\hat{\vvec}_i^\top (\vvec_j \sigma_j^{p-1} - \hat{\vvec}_j \sighat_j^{p-1})} + \sum_{i=\rdim+1}^\xdim \sqnorm{\hat{\vvec}_i^\top \hat{\vvec}_i \sighat_i^{p-1}} \\
&\le \sum_{i=\rdim+1}^\xdim \sum_{j=1}^\rankA \sqnorm{\vvec_j \sigma_j^{p-1} - \hat{\vvec}_j \sighat_j^{p-1}} + \sum_{i=\rdim+1}^\xdim  \sighat_i^{p-1} \\
&\le (\xdim -\rdim) \sum_{j=1}^\rankA \sqnorm{\vvec_j \sigma_j^{p-1} - \hat{\vvec}_j \sighat_j^{p-1}} + (d-r)  \sighat_\rdim^{p-1}\\
&= (\xdim -\rdim) \left[ \sum_{j=1}^\rankA \sqnorm{\vvec_j \sigma_j^{p-1} - \hat{\vvec}_j \sighat_j^{p-1}} + \sighat_\rdim^{p-1} - \sigma_\rdim^{p-1} \right] + (\xdim -\rdim)   \sigma_\rdim^{p-1}
\end{align*}
%
Combining this with the above, and taking the minimum of the two upper bounds, we get
\begin{align*}
\sqnorm{ \sum_{i=\rdim+1}^\xdim \hat{\vvec}_i \hat{\vvec}_i^\top\wvec^* }^2
&\le (\xdim - \rdim)\sigma_\rdim^{\picardpower-1} + (\xdim-\rdim) \min \left(
   \rankA  \sigma_{1}^{\picardpower-1}, \sum_{j=1}^\rankA \sqnorm{\vvec_j \sigma_j^{p-1} - \hat{\vvec}_j \sighat_j^{p-1}} + \sighat_\rdim^{p-1} - \sigma_\rdim^{p-1}\right)\\
   &= (\xdim - \rdim)\sigma_\rdim^{\picardpower-1} + (\xdim-\rdim) \epsilon(t)
\end{align*}
completing the proof.

\end{proof}

\section{Implementation details}\label{app:algorithm}

\subsection{Experimental set-up}
All experiments were run on a 12 core Intel Xeon E5-1650 CPU at 3.50GHz with 32 GiB of RAM. Each trial ran in its own thread with no more than 12 running in parallel. All algorithms were implemented in python using Numpy's and SciPy's math libraries. All matrix and vector operations were performed through Numpy's optimized subroutines. 

Two implementations of LSTD were used. The first builds the $\Amat$ matrix with incremental additions of outer-products while the second batches the operation in a faster matrix-matrix multiplication. 
Note that the batch version only allows $\lambda=0$. 
When solving the least-squares problem for LSTD, in both the sparse and dense case, either Numpy or SciPy's subroutines were used. In the dense case, to the best of our knowledge, Numpy used the same SVD subroutine as the one used for \tLSTD.
Our implementation of \tLSTD\ did not support sparse vectors but both TD and LSTD did.

For the runtime experiment, LSTD used the faster batch implementation. We made sure that enough memory was available for each thread such that no memory was required to be stored on disk. This was to ensure minimal interaction between the different threads. Note that  the cost of generating the data was not counted in the runtime of each algorithm. 

\subsection{Mini-batch algorithm}

We can perform the SVD-update 
in mini-batches of size $\rdim$
to obtain a computational complexity of $O(\xdim \rdim)$.
This update is given in Algorithm \ref{alg_tlstd_mini}.
The computational complexity is $O(\xdim \rdim^2 + \rdim^3)$ for
one call to the algorithm, but it is only called every $\rdim$ steps,
giving an amortized complexity of $O(\xdim \rdim)$ since $\xdim > \rdim$.

\begin{algorithm}[htp]
\caption{\textbf{update-svd}($\Umat,\Sigmamat,\Vmat, \Zmat, \Dmat, \rdim$) with mini-batches for \tLSTD}\label{alg_tlstd_mini}
\begin{algorithmic}[1]
  \State $\Qmat_Z, \Rmat_Z = \text{QR-decomposition}((\eye - \Umat\Umat^\top)\Zmat)$  \ \ \ \ \ \  \ \ \ // O($\xdim \rdim^2$) time by multiplying $\Umat^\top \Zmat$ first
    \State $\Qmat_D, \Rmat_D = \text{QR-decomposition}((\eye - \Vmat\Vmat^\top)\Dmat)$  \ \ \ \ \ \  \ \ \ // O($\xdim \rdim^2$) time
\State $ \Kmat \gets \left[ \begin{array}{cc}
\Sigmamat & \zerovec \\
\zerovec & \zerovec\end{array} \right]
+
\left[ \begin{array}{c}
\Umat^\top \Zmat \\
\Rmat_Z \end{array} \right]
\left[ \begin{array}{c}
\Vmat^\top \Dmat \\
\Rmat_D \end{array} \right]^\top
$
  \State $[\Lmat, \Sigmamat, \Rmat] \gets$ SVD($\Kmat$)  \ \ \ \ \ \  \ \ \ // O($\rdim^3$) 
  \State $\Umat \gets [\Umat \ \ \Qmat_Z] \Lmat$    \ \ \ \ \ \  \ \ \ // O($\xdim \rdim^2$) 
  \State $\Vmat \gets  [\Vmat \ \ \Qmat_D] \Rmat$   \ \ \ \ \ \  \ \ \ // O($\xdim \rdim^2$) 
  \Return $\Umat, \Sigmamat, \Vmat$
\end{algorithmic}
\end{algorithm}

\begin{algorithm}[htp!]
\caption{update-svd($\Umat, \Sigmamat, \Vmat, \Lmat, \Rmat, \zvec, \dvec, \rdim$) with one sample for incremental \tLSTD}\label{alg_tlstd_inc}
\begin{algorithmic}[1]
  \State $\mvec = \Lmat^\top \Umat^\top \zvec$ \ \ \ \ \ \  \ \ \ // O($\xdim \rdim$) time, as $\vvec = \Umat^\top \zvec$ is O($\xdim \rdim$) and $\Lmat^\top \vvec$ is O($\rdim^2$) 
  \State $\pvec = \zvec -  \Umat \Lmat \mvec$  \ \ \ \ \ \  \ \ \ // O($\xdim \rdim$) time
  \State $\nvec = \Rmat^\top \Vmat^\top \dvec$ \ \ \ \ \ \  \ \ \ // O($\xdim \rdim$) time
  \State $\qvec = \dvec - \Vmat \Rmat \nvec$ \ \ \ \ \ \  \ \ \ // O($\xdim \rdim$) time
  
\State $\Kmat \gets \left[ \begin{array}{cc}
\Sigmamat & \zerovec \\
\zerovec & 0\end{array} \right]
+
\left[ \begin{array}{c}
\mvec \\
\|\pvec\| \end{array} \right]
\left[ \begin{array}{c}
\nvec \\
\|\qvec\| \end{array} \right]^\top$
  \State $[\Lmatnext, \Sigmamat, \Rmatnext] \gets$ SVD($\Kmat$)
  \State $\Lmat \gets  \left[ \begin{array}{cc}
\Lmat & \zerovec \\
\zerovec & 1\end{array} \right]\Lmatnext$   \ \ \ \ \ \  \ \ \ // O($\rdim^3$) time
  \State $\Rmat \gets  \left[ \begin{array}{cc}
\Rmat & \zerovec \\
\zerovec & 1\end{array} \right]\Rmatnext$   \ \ \ \ \ \  \ \ \ // O($\rdim^3$) time

\If{$\| \pvec \| \le \epsilon$} 
\State $\| \pvec \| \gets \zerovec$ \ \ \ \ \ // $\epsilon = 0.00001$
\Else  \ $\pvec \gets \pvec / \| \pvec \| $ \ \ \ \ \ \  \ \ \ // normalize, update to $\Umat$ 
\EndIf
\If{$\| \qvec \| \le \epsilon$}
\State $\| \qvec \| \gets \zerovec$ \ \ \ \ \ // $\epsilon = 0.00001$
\Else \ $\qvec \gets \qvec / \| \qvec \| $ \ \ \ \ \ \  \ \ \ // normalize, update to $\Vmat$
\EndIf

\State $\Umat \gets [\Umat \ \ \pvec]$   \ \ \ \ \ \  \ \ \ // Only allowed to grow to $2\rdim$ columns
\State $\Vmat \gets [\Vmat \ \ \qvec]$   \ \ \ \ \ \  \ \ \ // Only allowed to grow to $2\rdim$ columns

\State // If reached size $2 \rdim$, reduce back to $\rdim$ by dropping smallest $\rdim$ singular values; O($\xdim\rdim$) amortized complexity
\If{size($\Lmat) \ge 2\rdim$} 
  \State $\Sigmamat \gets \Sigmamat(1:\rdim,1:\rdim)$
  \State $\Umat \gets \Umat \Lmat$  \ \ \ \ \ \  \ \ \ // O($\xdim\rdim^2$) time
  \State $\Umat \gets \Umat(:,1:\rdim)$   \ \ \ \ \ \  \ \ \ // Concatenate back to $\rdim$ columns 
  \State $\Vmat \gets \Vmat \Rmat$  \ \ \ \ \ \  \ \ \ // O($\xdim\rdim^2$) time
  \State $\Vmat \gets \Vmat(:,1:\rdim)$   \ \ \ \ \ \  \ \ \ // Concatenate back to $\rdim$ columns    
    \State $\Lmat = \eye$, $\Rmat = \eye$  \ \ \ \ \ \  \ \ \ // Reintialize   
\EndIf   
\\ \Return $\Umat, \Sigmamat, \Vmat, \Lmat, \Rmat$
\end{algorithmic}
\end{algorithm}

\begin{algorithm}[htp]
\caption{compute-weights($\Umat, \Sigmamat, \Vmat, \Lmat, \Rmat, \bvec$), O($\xdim \rdim$)}\label{alg_w}
\begin{algorithmic}[1]
\State // Solve $\Amat^\inv \bvec$ where $\Amat = \Umat \Lmat \Sigmamat \Rmat^\top \Vmat^\top$ and so $\Amat^\inv = \Vmat \Rmat\Sigmamat^\inv \Lmat^\top \Umat^\top$
\State // Does not invert any singular values that are below $0.01*\sighat_1$
\State $\tilde{\bvec} = \Lmat^\top \Umat^\top \bvec$  \ \ \ \ \ \  \ \ \ // O($\xdim \rdim$) time, implicit left singular vector is $\Umat \Lmat$
\State $\sighat_1 \gets \Sigmamat(1,1)$
\State $\Sigmamat^\pinv \gets \zerovec$ \ \ \ \ \ \  \ \ \ // initialize as zero matrix
\For{$i \in \{1, \ldots, \rdim\}$}
\If{$\Sigmamat(i,i) > 0.01 \sighat_1$}
\State $\Sigmamat^\pinv(i,i) \gets  \Sigmamat(i,i)^{-1}$
\Else
\State \textbf{break}
\EndIf
\EndFor
  \State $\wvec = \Vmat \Rmat \Sigmamat^\inv \tilde{\bvec}$ \ \ \ \ \ \  \ \ \ // O($\xdim \rdim$) time
\end{algorithmic}
\end{algorithm}

\subsection{Fully incremental algorithm}
To perform an update and compute the
current function approximation solution $\wvec$ on each time
step, we can no longer amortize all costs across
$\rdim$ steps. In this setting, however, we can
obtain some nice efficiency improvements by 
exploiting the form of the update for one sample
to obtain an $O(\xdim \rdim + \rdim^3)$ algorithm. 
We opt for the simplest implementation in this work,
using a singular value decomposition to diagonalize $\Kmat$.
We could improve efficiency by
using the Lanczos bi-diagonalization algorithm with full orthogonalization \citep{golub1965calculating,roman2008arobust};
however, we leave these additional speed improvement
for future work, and focus on a more vanilla \tLSTD\ implementation
in this work. 
Note that to avoid $O(\xdim \rdim^2)$ computation to update $\Umat = \Umat \Lmat$,
the matrix $\Lmat$ is saved for the next iteration and first applied to the vectors.

An additional aspect of the algorithm is to enable the size of the approximation to grow
to $2\rdim$. For the fully incremental setting, new information may be too quickly thrown away
if truncation is performed on each step. By allowing the subspace to grow, the algorithm
should better track and incorporate new information; after $2\rdim$ steps, it can then truncate. 
This does not change the computational complexity of the algorithms, in terms of its order,
because the runtime is only multiplied by these constants. Further, to ensure that we maintain O($\xdim \rdim$)
for the matrix multiplications, we perform the $\Umat \Lmat$ computation in this step,
to amortize the $O(\xdim \rdim^2)$. Periodically performing this computation
is important for numerical stability, though it remains future work to more fully
understand how frequently this should be done, as well as how much the subspace should be
allowed to grow. 
 
\subsection{Competitor algorithms}\label{sec_competitors}


We highlight some algorithms that we did not compare to,
and carefully justify why they do not match the setting
for which \tLSTD\ is designed. 
LSTD was not included in the comparison on the energy domain,
because it is computationally infeasible for $\xdim = 40,000$,
having a storage and computational complexity of $O(\xdim^2) = 1.6$ billion. 
We did not compare to fLSTD-SA \cite{prashanth2013fast},
since that algorithm is not designed for the streaming setting, but rather
requires a batch of data upfront from which it can randomly subsample.
In fact, it is much more similar to a TD algorithm, but where
samples are drawn uniformly randomly.
Similarly, random projections for LSTD \cite{ghavamzadeh2010lstd}
was introduced and analyzed for the batch setting.
Finally, forgetful LSTD is also designed for a different purpose \cite{vanseijen2015adeeper},
with the goal to improve upon LSTD and linear Dyna. Consequently, 
the focus of the algorithm is not computational efficiency, and it is at least O($\xdim^2$)
in terms of memory and storage. 

%

\section{Properties of $\Amat$ in benchmark tasks}\label{app_sing}

For these experiments, a total of 10000 episodes with random start states, were used to obtain the empirical $\Amat$.

\begin{figure*}
	\subfigure[Mountain Car w/ Tiles]{\includegraphics[width=\figwidthfour]{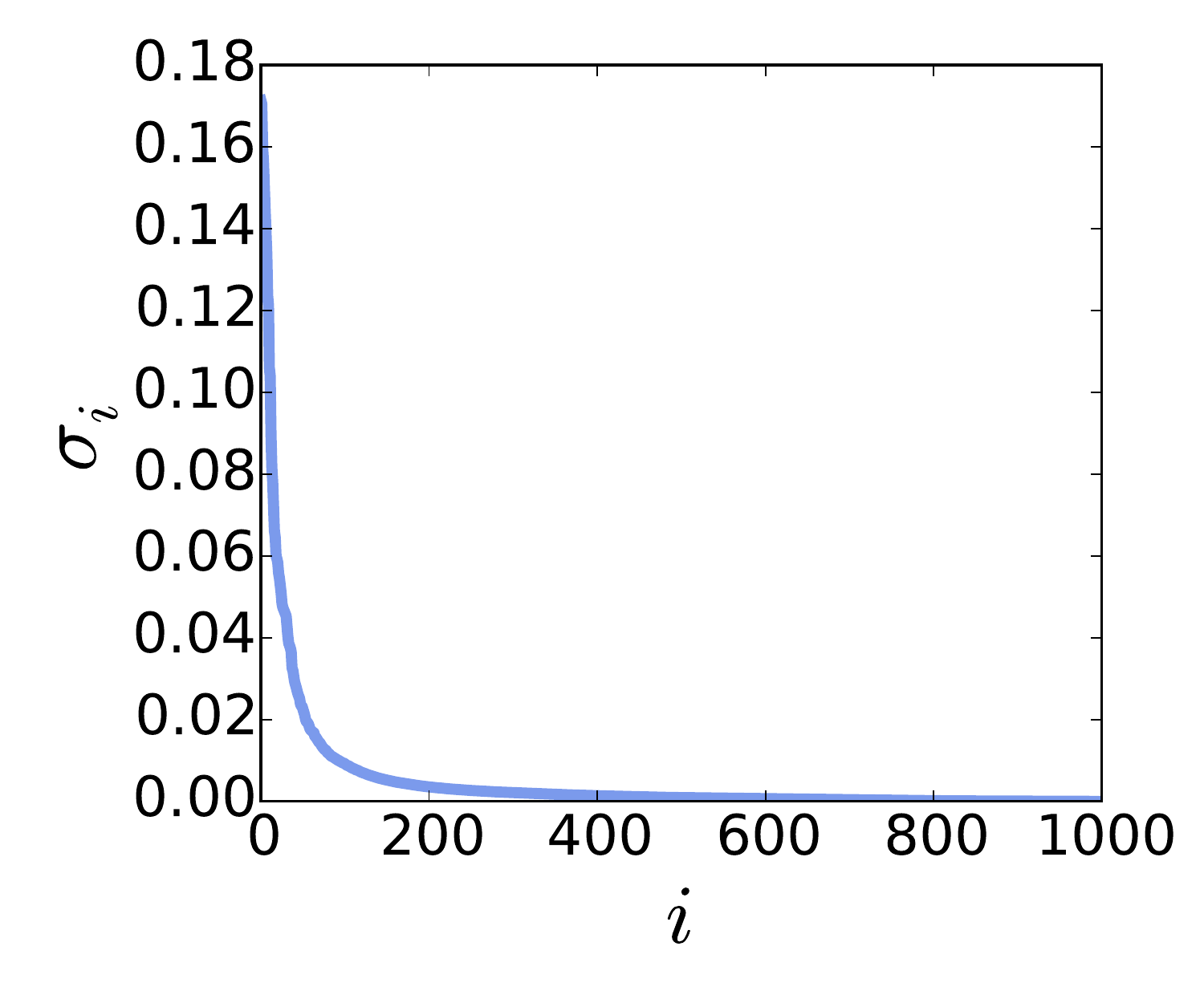}
		\label{fig:tile-sig-val-mc}}
	\subfigure[Mountain Car w/ RBFs]{
		\includegraphics[width=\figwidthfour]{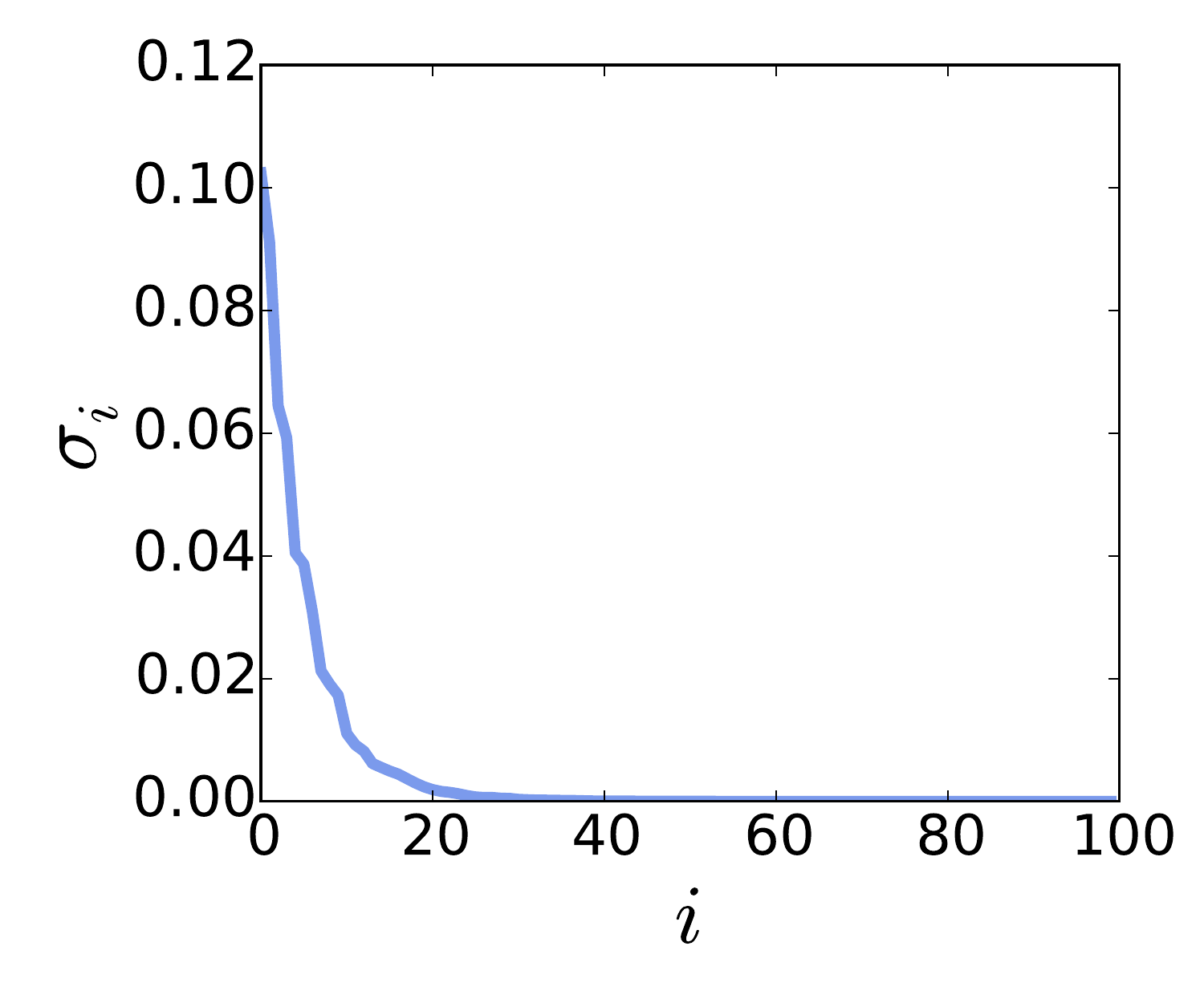}
		\label{fig:rbf-sig-val-mc}}
	\subfigure[Mountain Car w/ Tiles]{
		\includegraphics[width=\figwidthfour]{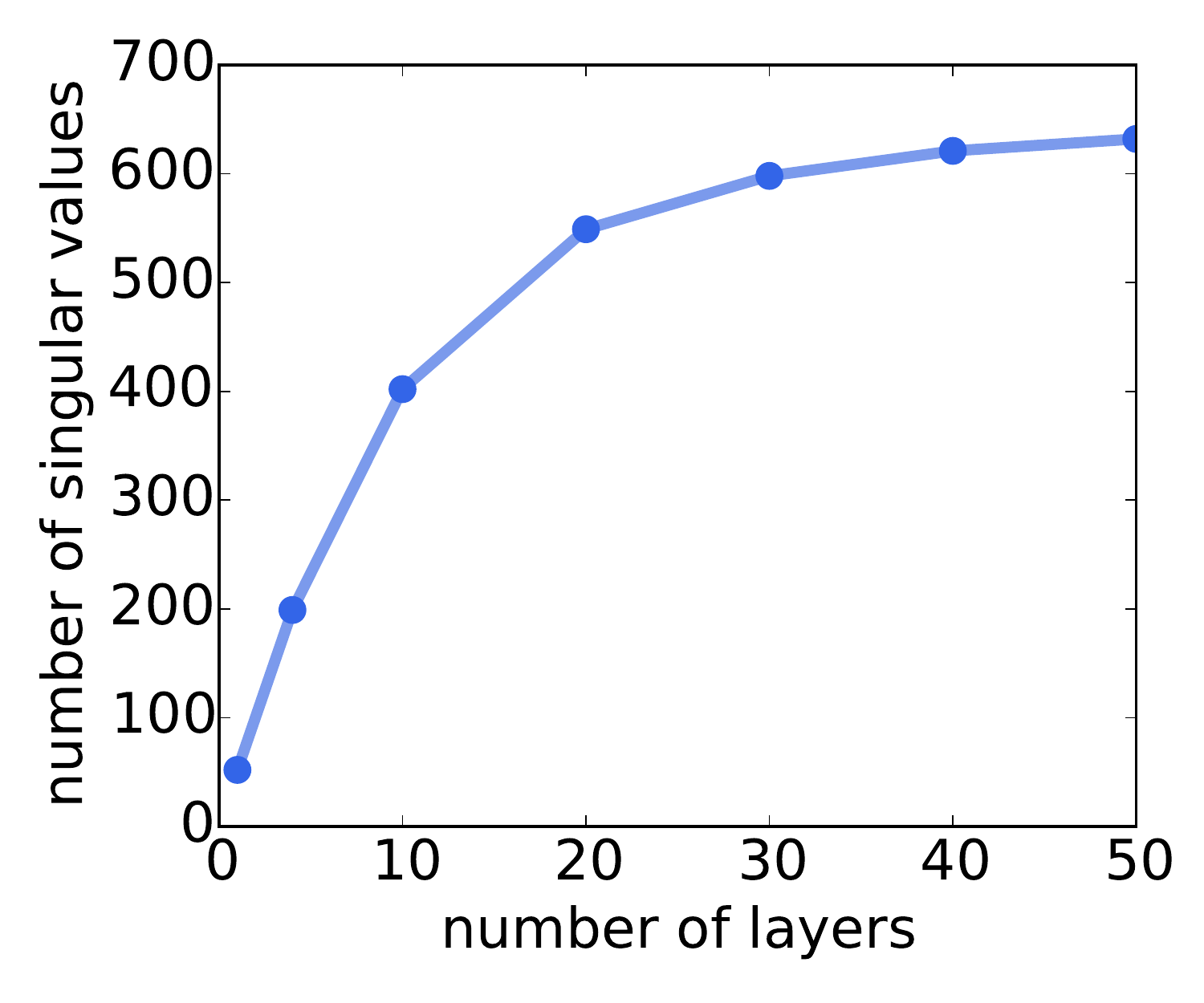}
		\label{fig:tile-sig-num-mc}}
	\subfigure[Mountain Car w/ RBFs]{
		\includegraphics[width=\figwidthfour]{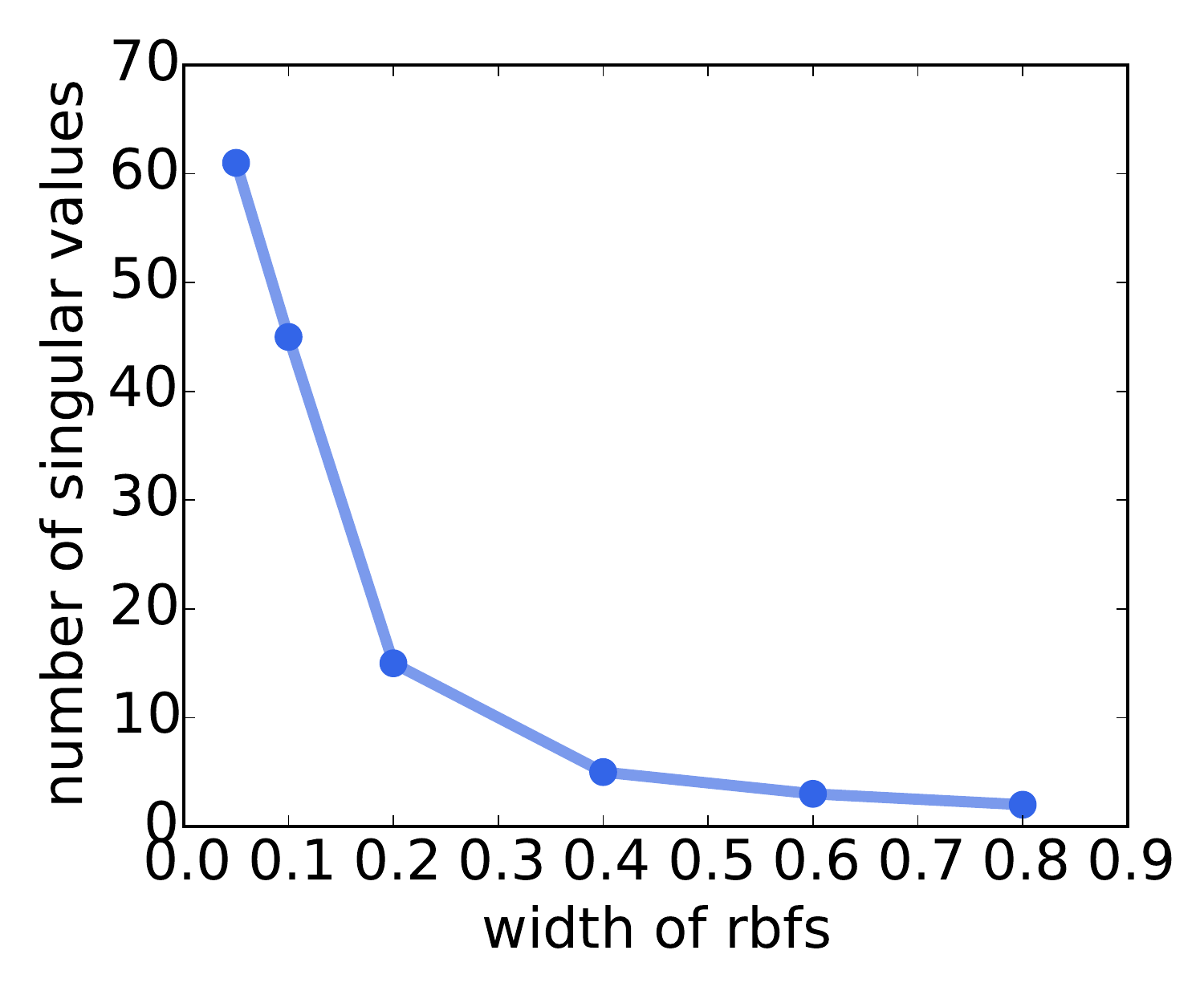}
		\label{fig:rbf-sig-num-mc}}\\\\
	\subfigure[Pendulum w/ Tiles]{
		\includegraphics[width=\figwidthfour]{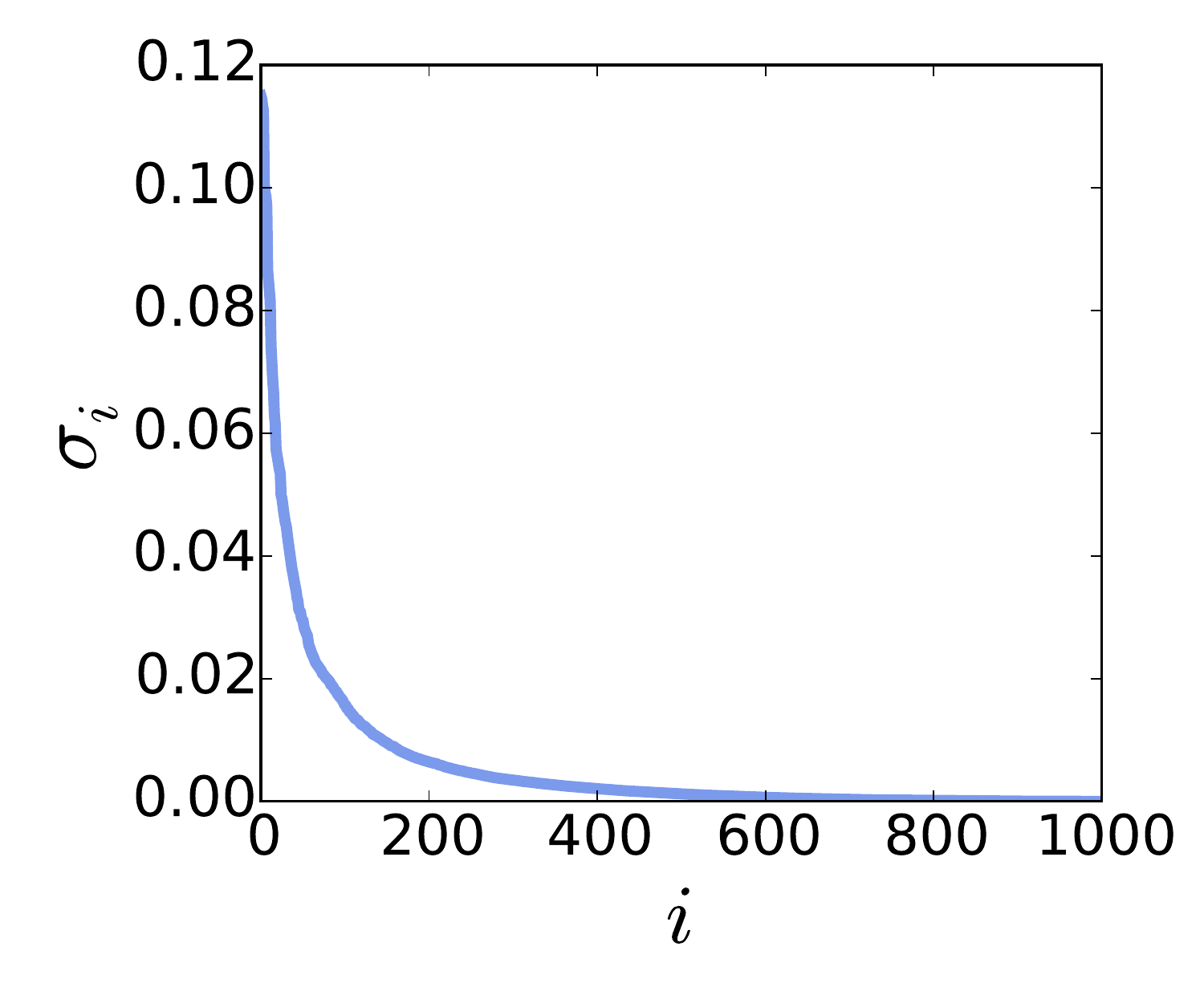}
		\label{fig:tile-sig-val-p}}
	\subfigure[Pendulum w/ RBFs]{
		\includegraphics[width=\figwidthfour]{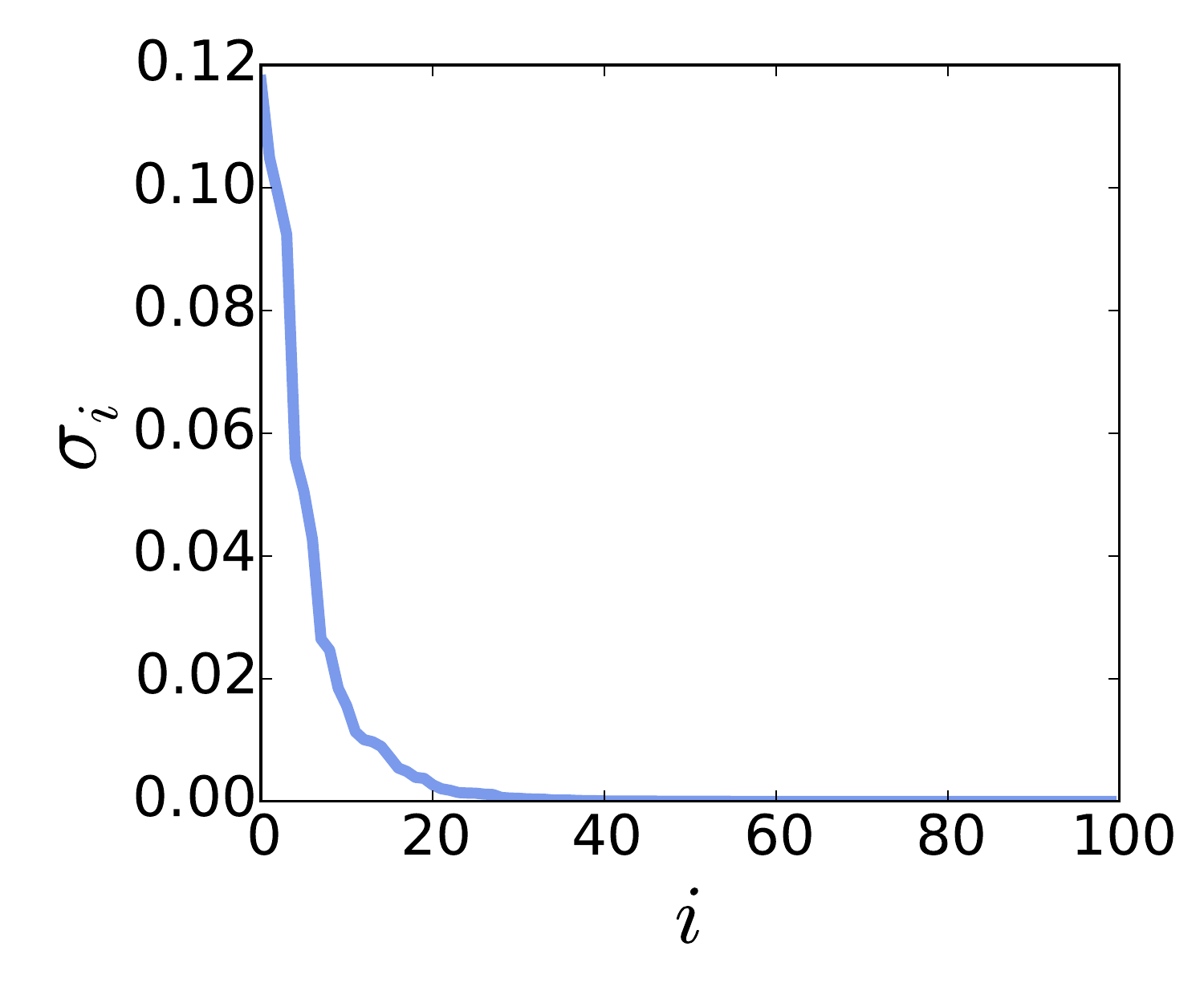}
		\label{fig:rbf-sig-val-p}}
	\subfigure[Pendulum w/ Tiles]{
		\includegraphics[width=\figwidthfour]{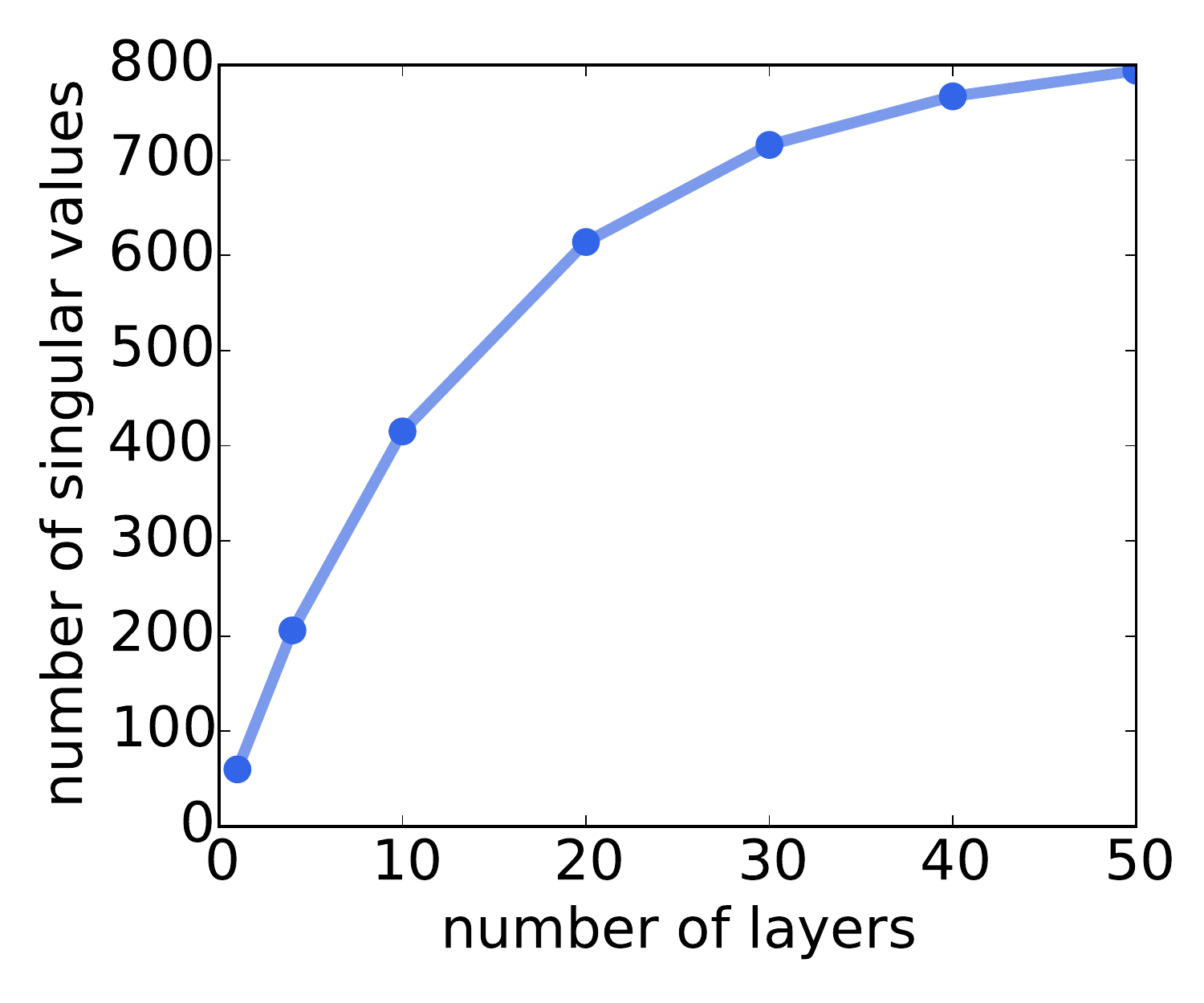}
		\label{fig:tile-sig-num-p}}
	\subfigure[Pendulum w/ RBFs]{
		\includegraphics[width=\figwidthfour]{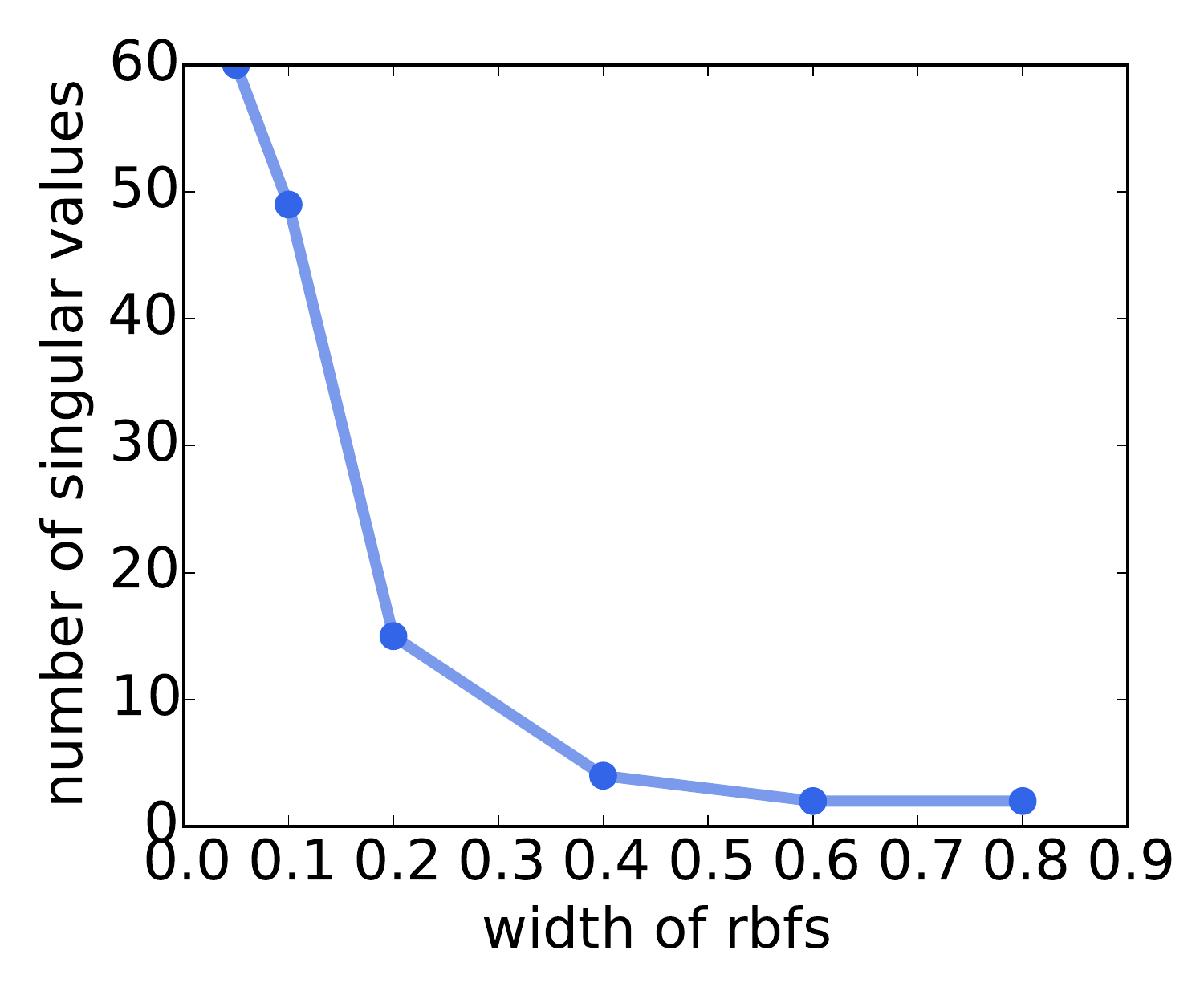}
		\label{fig:rbf-sig-num-p}}
	\caption{
	\textbf{(a) and (e)} The sorted singular values of $\Amat$ with tile coding using 10 layers of 10x10 grids.
	\textbf{(b) and (f)} The sorted singular values of $\Amat$ with tile coding using a 10x10 grids of RBFs with widths equal to $0.2 \times$ the range of the state space.
	\textbf{(c) and (g)} Number of singular values required to have $95\%$ of the total weight of the singular values of $\Amat$ with tile coding using varying number of layers of 10x10 grids.
	\textbf{(d) and (h)} Number of singular values required to have $95\%$ of the total weight of the singular values of $\Amat$ using a 10x10 grid of RBFs with varying widths reported as some fraction of the state space range.}
	\label{fig:allsing}
\end{figure*}

The number of singular values required to accurately represent $\Amat$ follows an interesting trend. First, in Figure \ref{fig:tile-sig-num-mc} and \ref{fig:tile-sig-num-p}, we see that as we change the number of features by adding layers to the tile coding, the required number of features plateaus. 
This opens up the possibility of using very rich representations with tile coding while keeping the representation of $\Amat$ compact. Secondly, in Figure \ref{fig:rbf-sig-num-mc} and \ref{fig:rbf-sig-num-p}, we observe a rapid drop in the number of singular value required as the width of the RBFs increases. This shows that this form of approximation for $\Amat$ can effectively leverage dependencies in the features, potentially giving a designer more flexibility in choosing which features to include, while letting the algorithm extract the relevant information.

\section{Additional value function accuracy graphs}\label{app_benchmark}

The learning curves for the benchmark domains in the main paper
are only for Mountain Car with RBFs. The below includes results
for tile coding, and further for Pendulum. These experiments, combined with
the singular values graphs, highlight that \tLSTD\ is well-suited for
problems where $\rdim$ can be set large enough to incorporate the majority
of the large singular values. The singular values for the tile coding representation 
in the benchmark tasks did not drop as quickly as for RBFs; 
this is clearly reflected in the performance of \tLSTD. In fact, we do not expect
\tLSTD\ to perform well in settings where $\Amat$ has more than $\rdim$ large singular values. 
Interestingly, however, even with $\rdim$ smaller than needed for a system, \tLSTD\
still obtains early learning gains. This suggests interesting avenues moving forward,
for combining \tLSTD\ and TD, or other strategies for robustness when \tLSTD\
is applied to systems where the $\Amat$ matrix has more than $\rdim$ large singular values.
The below includes the other graphs already shown in the paper, to make it easier
to look at the results together. 

\begin{figure*}[htp!]
\centering
	\subfigure[Mountain Car w/ Tile coding]{
		\includegraphics[width=\figwidthtwo]{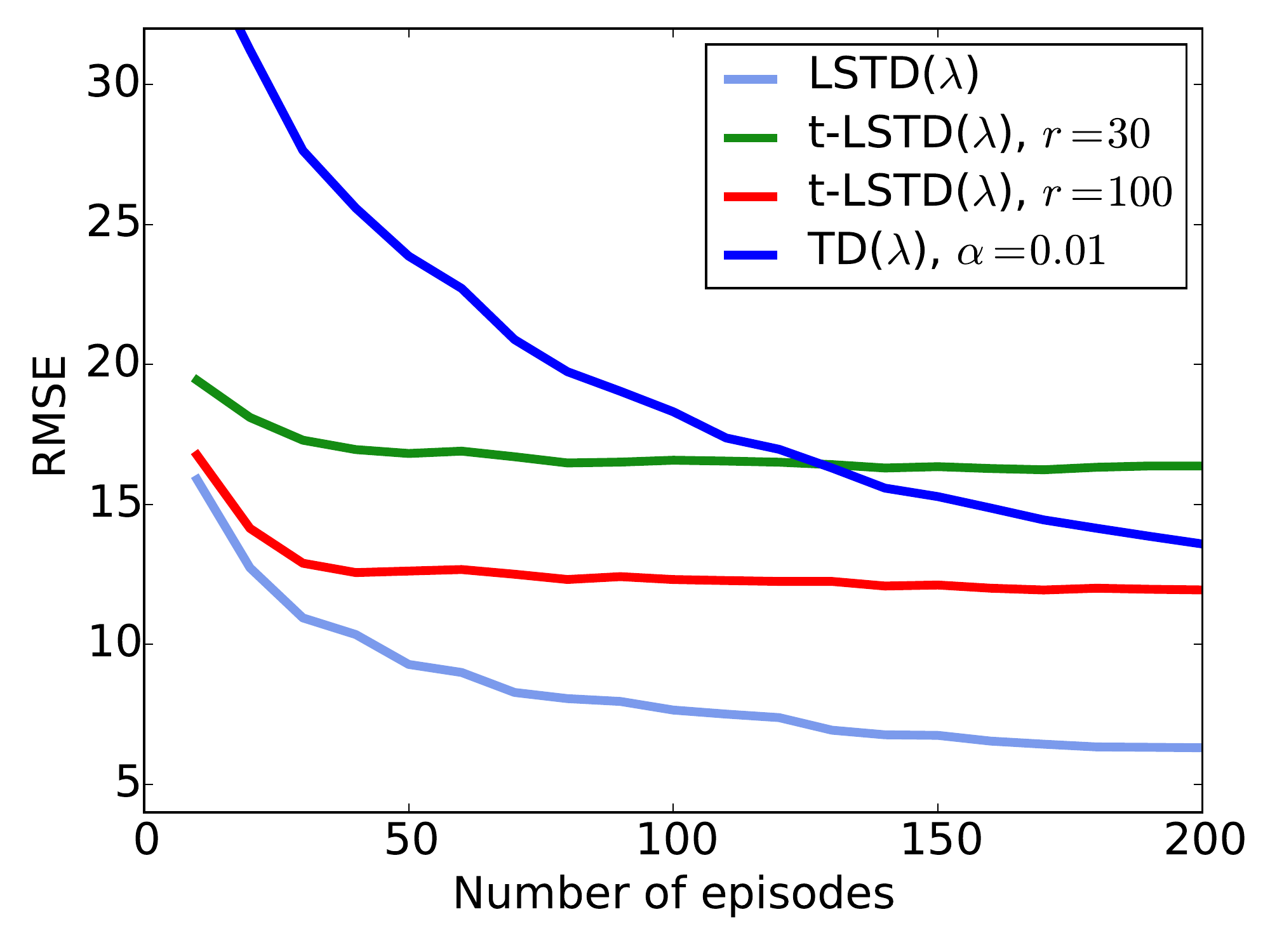}\label{fig:tile-compare-mc-complete}}
	~
	\subfigure[Pendulum w/ Tile coding]{
		\includegraphics[width=\figwidthtwo]{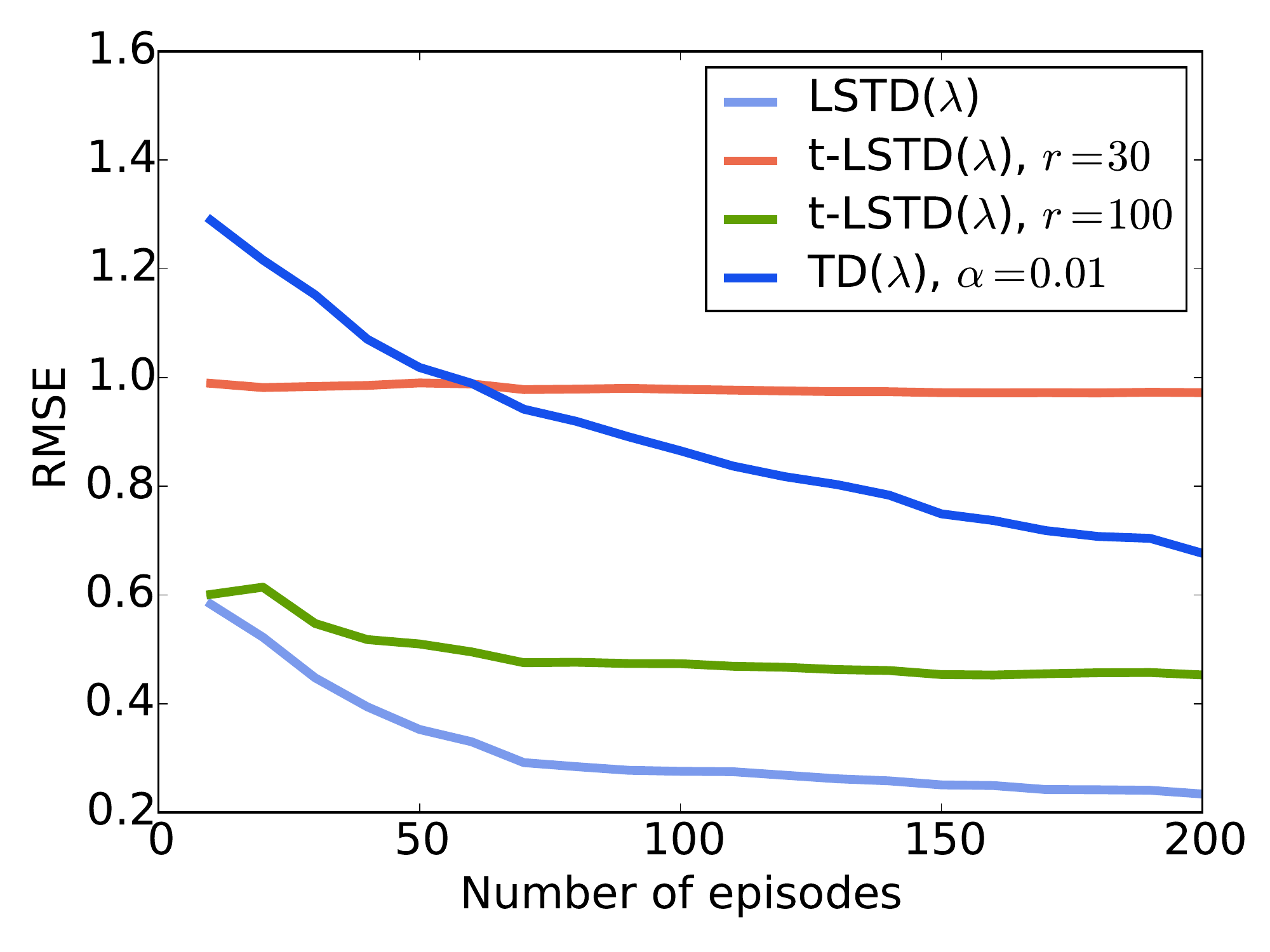} \label{fig:tile-compare-p-complete}}\\	
	\subfigure[Mountain Car w/ RBFs]{
		\includegraphics[width=\figwidthtwo]{figures/rbf_mc_final.pdf}\label{fig:rbf-compare-mc-complete}}
	~
	\subfigure[Pendulum w/ RBFs]{
		\includegraphics[width=\figwidthtwo]{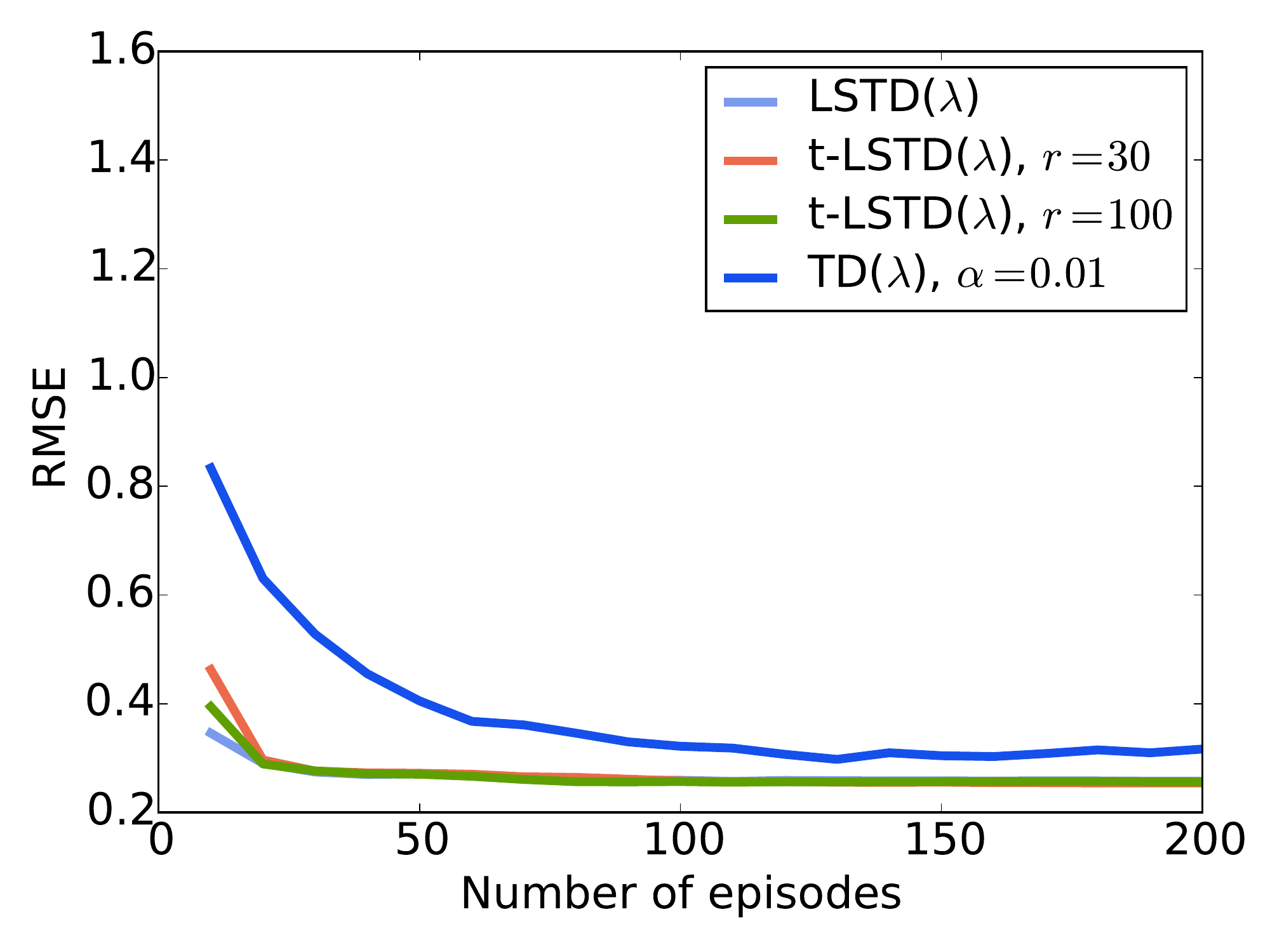}\label{fig:rbf-compare-p-complete}}	\\
	~
	\subfigure[Rank versus performance with tile coding in Mountain Car]{
	         \includegraphics[width=\figwidthtwo]{figures/tile_mc_rankvsperf.pdf}\label{fig:rank_tile-complete}}
	~
	\subfigure[Rank versus performance with RBFs in Mountain Car]{
		\includegraphics[width=\figwidthtwo]{figures/rbf_mc_rankvsperf.pdf}\label{fig:rank_rbf-complete}}			
	\caption{Root Mean Squared Error (RMSE) of the true discounted returns and the learned value function
	for several different scenarios. The RMSE is reported in the two domains 
	\textbf{For (a) and (b)} for tile coding using 10 layers of 10x10 grids.
	\textbf{For (c) and (d)} for a grid of 32x32 RBFs with width equal to $0.12$ times the total range of the state space.
	\textbf{For (e) and (f)} versus the chosen rank $\rdim$. We can see that
	large $\rdim$ are not necessary, with performance leveling off at $\rdim = 50$. For high values of $\rdim$ and fewer samples, the error slightly increases, likely due to some instability with incremental updating and very small singular values. 
	}
	\label{fig:rmse-compare-complete}
\end{figure*}

\begin{figure}[htp!]
\centering
		\includegraphics[width=\figwidthtwo]{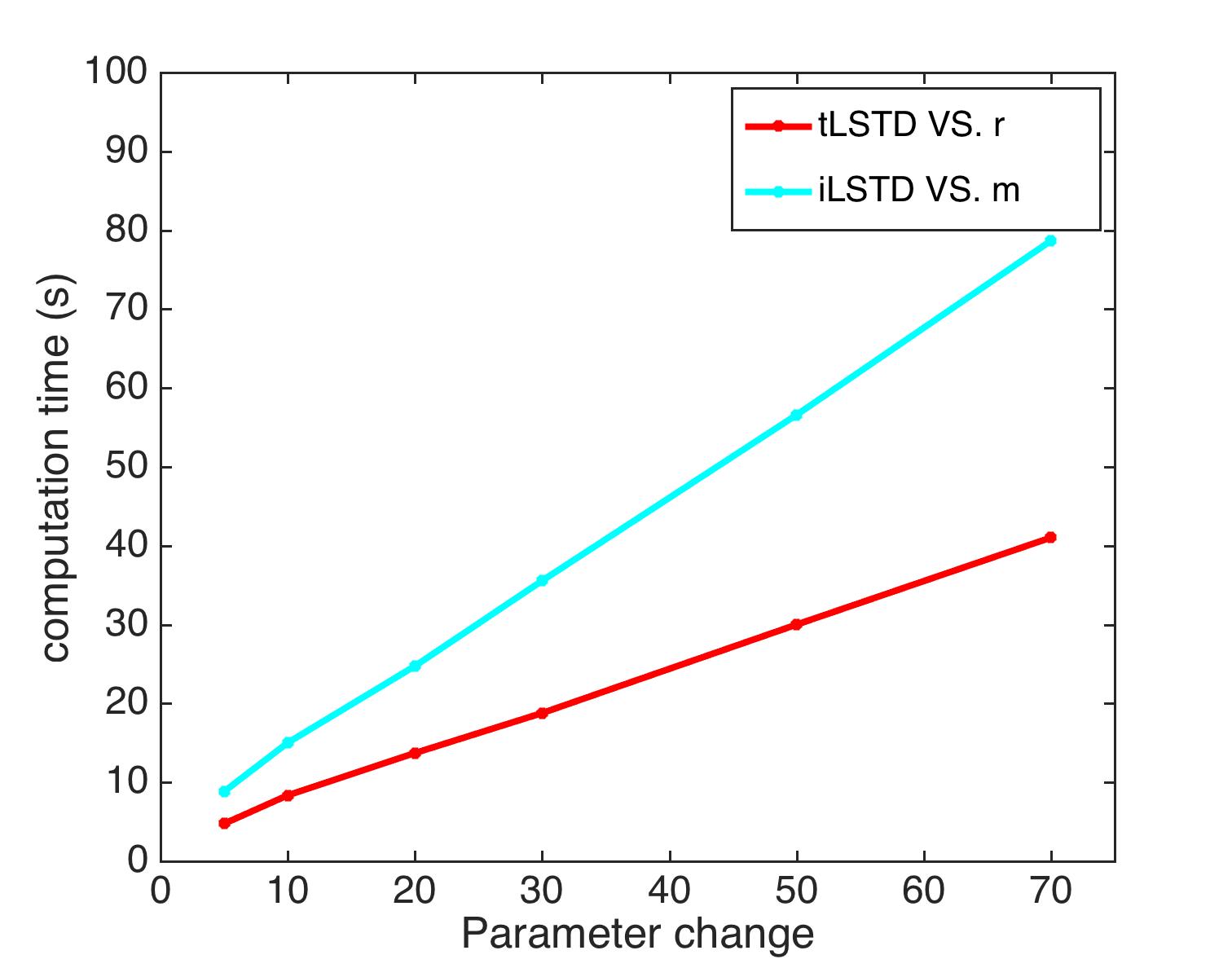}\label{fig:time_rank_m}
\caption{Runtime with increasing $r$ or $m$, where $\rdim$ is the input rank for \tLSTD\ algorithm and $m$ is the number of parameters to update at each step in iLSTD algorithm.  Runtimes are averaged over 30 trajectories of length 500. 
}
\end{figure}

Finally, Figure \ref{fig:time_rank_m} demonstrates runtimes for \tLSTD\ and iLSTD, for increasing parameters
$\rdim$ and $m$. These are corresponding parameters in the two algorithms in
that both result in O($\xdim \rdim$) and O($\xdim m$) runtime, respectively.
Despite this equivalence in terms of order, we see that \tLSTD actually scales better with $\rdim$.
This may be because iLSTD stores a matrix of size $\xdim^2$, and accesses columns of it $m$ times.
Several steps in iLSTD are implemented with sparse operations, but the matrix $\Amat$ itself is not sparse.

\end{document}